%% file: acl_latex.tex
\pdfoutput=1
\documentclass[11pt]{article}

\usepackage[final]{acl}

\usepackage{times}
\usepackage{latexsym}
\usepackage{inconsolata}
\usepackage{booktabs}
\usepackage{multirow}
\usepackage{graphicx}
\usepackage{amssymb}
\usepackage{mathrsfs}
\usepackage{amsmath}
\usepackage{bbding}
\usepackage{colortbl}
\usepackage{tcolorbox}
\usepackage{amsthm}
\usepackage{url}
\usepackage{wrapfig}
\usepackage{enumitem}

\usepackage[linesnumbered, ruled, vlined]{algorithm2e}
\usepackage{float}
\usepackage{subcaption}
\usepackage{amsfonts}
\usepackage{nicefrac}
\usepackage{microtype}
\usepackage{xcolor}
\usepackage{multirow}

\usepackage[T1]{fontenc}
\usepackage[utf8]{inputenc}

\usepackage{microtype}

\usepackage{inconsolata}

\usepackage{graphicx}

\title{Data Mixing Agent: Learning to Re-weight Domains for Continual Pre-training}

\author{Kailai Yang\textsuperscript{1}\textsuperscript{*}\quad Xiao Liu\textsuperscript{2}\textsuperscript{\dag}
\quad 
\textbf{Lei Ji}\textsuperscript{\textbf{2}}\quad \textbf{Hao Li}\textsuperscript{\textbf{3}} \quad \textbf{Xiao Liang}\textsuperscript{\textbf{4}}\quad \textbf{Zhiwei Liu}\textsuperscript{\textbf{1}} \\ \textbf{Yeyun Gong}\textsuperscript{\textbf{2}}\textsuperscript{\dag} \quad \textbf{Peng Cheng}\textsuperscript{\textbf{2}}\quad\textbf{Mao Yang}\textsuperscript{\textbf{2}}\\
    \textsuperscript{1} The University of Manchester\\
    \textsuperscript{2} Microsoft Research\\ 
    \textsuperscript{3} Imperial College London\\
    \textsuperscript{4} University of California, Los Angeles \\
}

\begin{document}

\maketitle
\begin{abstract}
Continual pre-training on small-scale task-specific data is an effective method for improving large language models in new target fields, yet it risks catastrophic forgetting of their original capabilities. A common solution is to re-weight training data mixtures from source and target fields on a domain space to achieve balanced performance. Previous domain reweighting strategies rely on manual designation with certain heuristics based on human intuition or empirical results. In this work, we prove that more general heuristics can be parameterized by proposing \textbf{Data Mixing Agent}, the first model-based, end-to-end framework that learns to re-weight domains. The agent learns generalizable heuristics through reinforcement learning on large quantities of data mixing trajectories with corresponding feedback from an evaluation environment. Experiments in continual pre-training on math reasoning show that Data Mixing Agent outperforms strong baselines in achieving balanced performance across source and target field benchmarks. Furthermore, it generalizes well across unseen source fields, target models, and domain spaces without retraining. Direct application to the code generation field also indicates its adaptability across target domains. Further analysis shows the agents' well-aligned heuristics with human intuitions and efficiency in achieving superior model performance with less source-field data.
\end{abstract}

\renewcommand{\thefootnote}{\fnsymbol{footnote}}
\footnotetext[1]{Work done during his internship at Microsoft Research.}
\footnotetext[2]{Corresponding authors.}
\renewcommand{\thefootnote}{\arabic{footnote}}

\input{Sections/intro}
\input{Sections/task_formulation}
\input{Sections/method}
\input{Sections/experiments}
\input{Sections/related_work_simple}
\input{Sections/conclusion}
\input{Sections/limitations}

\bibliography{references}

\appendix

\input{Appendix/sections/heuristic}
\input{Appendix/sections/start_state_estimate}
\input{Appendix/sections/implementation}
\input{Appendix/sections/experiment_set}
\input{Appendix/sections/experiment_result}
\input{Sections/related_work}

\end{document}

%% file: Sections/intro.tex
\section{Introduction}
Large Language Models (LLMs)~\citep{yang2025qwen3,liu2024deepseek}, though pre-trained to obtain generalization capabilities, often require further enhancement in knowledge-intensive fields~\citep{yang2024qwen2,guo2024deepseek} via continual pre-training in the target field. 
However, directly adapting to the target field data can lead to catastrophic forgetting of source data and collapse on existing model capabilities~\citep{hui2024qwen2,lin2025sigma}, due to the significant distribution shift between source and target fields. 

A popular solution is to curate data mixtures of the source and target fields to achieve a balanced performance~\citep{shi2024continual}. Existing methods mainly organize data mixtures defined by meta-attributes such as data sources and focus, known as domains~\citep{du2022glam,luo2024velocitune}. During training, the data mixture is allocated through a distribution in the domain space, which reflects the ratio of data allocated from each domain.
The distribution can be adjusted after several training steps if necessary, leading to a data mixing trajectory~\citep{luo2024velocitune,xia2023sheared} along the domain reweighting steps. Data mixing trajectories significantly influence model performance~\citep{olmo20242,grattafiori2024llama,li2024datacomp}, and previous works have explored data mixing algorithms to determine optimal trajectories for different tasks~\citep{liu2024regmix,ye2024data,xie2023doremi,xia2023sheared,luo2024velocitune}.

\begin{figure*}[t]
\centering
\includegraphics[width=0.8\textwidth]{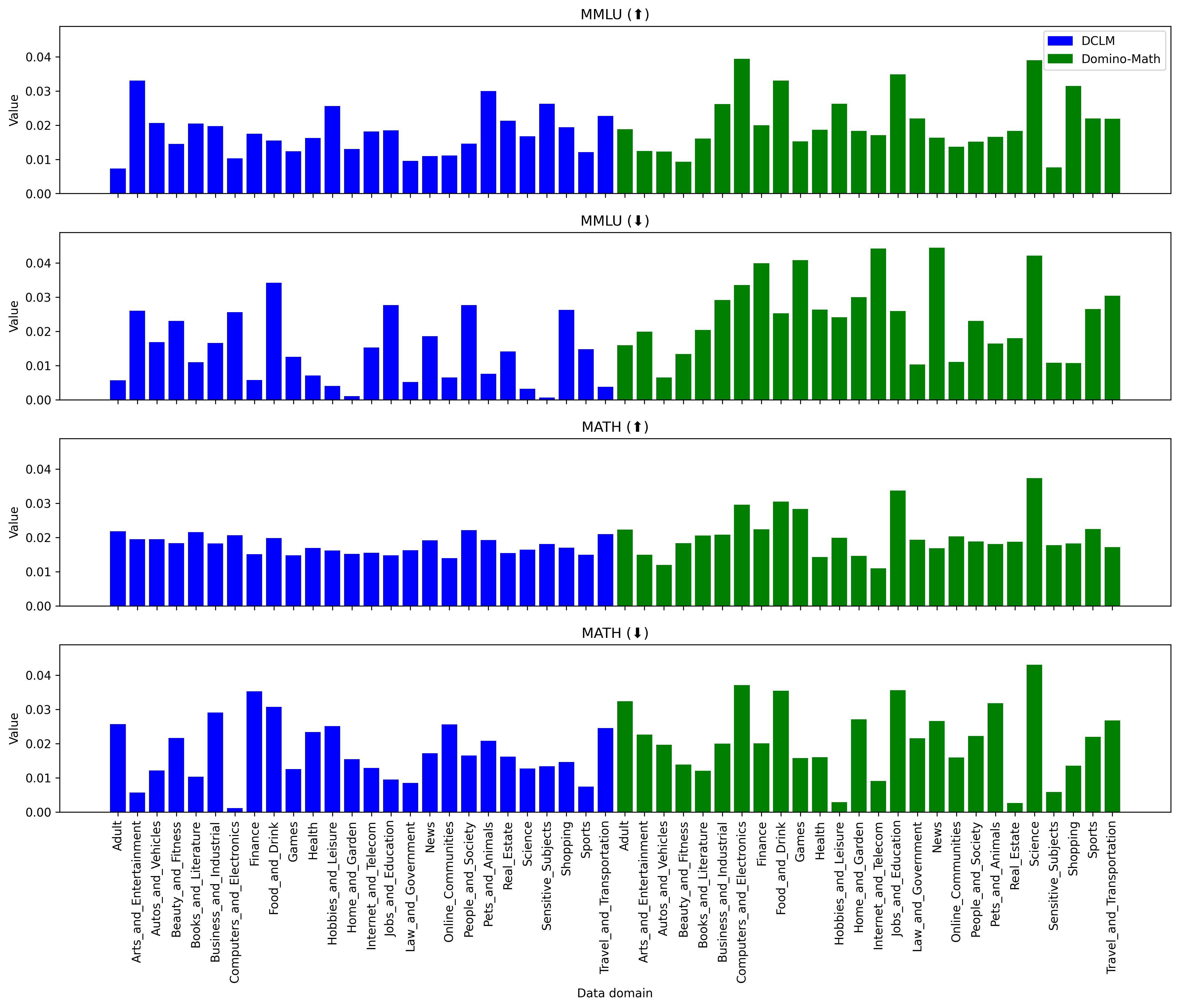}
\caption{Four averaged distributions drawn from 20 randomly generated data mixing trajectories. Each distribution in the trajectories is first categorized by whether it increases/decreases the performance of a 50M target model on the MMLU or MATH benchmarks within one re-weighting step, leading to four categories. The distributions in each category are then averaged within the domain space to represent their features, as shown in the figure. The models are trained on a 52-dimensional space, mixing the general domain data from DCLM~\citep{li2024datacomp} and the math reasoning data from the math split of the Dolmino-mix-1124~\citep{olmo20242} dataset.}
\label{fig:vertical_histograms_example}
\end{figure*}

A commonality of these data mixing methods is that their designs are based on certain general heuristics, such as: \textit{data mixtures that provide balanced evaluation loss lead to desired downstream performance}. Another indication of these heuristics is the various empirical conclusions drawn from training practices. For example, \citet{wettig2025organize} concluded that \textit{Data from the ‘Science' domain heavily promote model performance on MMLU~\citep{hendrycks2020measuring}, while the ‘Home' domain improves HellaSwag~\citep{zellers2019hellaswag} performance}. 
In Fig. \ref{fig:vertical_histograms_example}, we provide an average of distributions along 20 randomly generated data mixing trajectories (each with 80 domain reweighting steps), separated into four categories. The categories are defined by whether they increase or decrease the target model's performance on the MMLU/MATH~\citep{hendrycks2021measuring} benchmarks. 
The data mixing domains are defined and classified by the Nvidia domain classifier~\footnote{\url{https://huggingface.co/nvidia/domain-classifier}}. According to the results,
there are explicit differences between data distributions that increase/decrease model performance. For example, in MMLU (the first and second rows), higher ratios of DCLM data from the \textit{Science} and \textit{Home$\&$Garden} domains significantly improves benchmark performance. In MATH, increasing data mixtures from the \textit{Hobbies$\&$Leisure} and \textit{Real estate} domains of the  Dolmino-math data while keeping a balanced mix of DCLM data is likely to boost benchmark performance. 
The above results unveil more general heuristics that can enhance model performance in continual pretraining, yet have not been discovered or utilized by existing works. This negligence motivates further efforts to efficiently mine such heuristics and leverage them for data mixing in different scenarios.
% We conduct preliminary experiments on continual pretraining and discover more examples of such general heuristics in Appn. \ref{appn:general_heuristic}.

These observations reveal a rich heuristic space for domain reweighting. We believe these model- and data-agnostic heuristics can be unified into a small agent model to guide the data mixing trajectories in an end-to-end manner.
Based on this intuition, we propose \textbf{Data Mixing Agent}, the first model-based method that learns to re-weight domains for continual pre-training. We start by randomly sampling large quantities of data mixing trajectories, each with fixed domain re-weighting steps. We then train small proxy models on all trajectories, obtaining model checkpoints on each re-weighting step. All checkpoints are evaluated in a light-weight evaluation environment to assess the target capabilities. The trajectories and corresponding environment feedback are expected to empirically enclose a wide range of heuristics. Data Mixing Agent is then optimized on these collected data with the Conservative Q-Learning (CQL) reinforcement learning algorithm~\citep{kumar2020conservative}. During continual pre-training on the target model, the agent directly predicts the domain distribution for the next domain reweighting step on the fly, considering previous states in the data mixing trajectory and the environment feedback.

We apply Data Mixing Agent on the math reasoning target field while preserving performance in the general field. Evaluation on in-distribution source field data shows that our method significantly outperforms state-of-the-art static~\citep{liu2024regmix} and dynamic~\citep{xia2023sheared} domain reweighting methods, achieving an average improvement of 3.02\% across 8 general benchmarks and 4 math reasoning benchmarks. The agent's generalization ability is demonstrated with balanced performance across 3 unseen source fields, 4 target models, and 2 domain spaces, all without retraining. We also directly apply agents trained on math reasoning to guide training on the unseen code generation field, proving their generalization across target domains. Additional analysis confirms that these heuristics align well with human intuitions, and Data Mixing Agent can achieve superior continual pre-training performance with less source-field data.

In summary, we make the following contributions: 
\begin{itemize}
    \item We propose Data Mixing Agent, the first end-to-end, lightweight domain reweighting method for continual pre-training;
    \item Data Mixing Agent significantly outperforms state-of-the-art baselines in extensive experiments, leading to balanced performance across model capabilities and high efficiency in data usage;
    \item Data Mixing Agent learns heuristics that generalize across source and target fields, target models, and domain spaces.
\end{itemize}
% enabling direct application in multiple scenarios.

% In summary, this work makes the following contributions:
% \begin{itemize}
%     \item We propose Data Mixing Agent, the first end-to-end, lightweight domain reweighting method for continual pre-training;
%     \item Data Mixing Agent significantly outperforms state-of-the-art baselines in extensive experiments, leading to balanced performance across model capabilities and high efficiency in data usage;
%     \item Data Mixing Agent learns heuristics that generalize across source and target fields, target models, and domain spaces, enabling direct application in multiple scenarios.
% \end{itemize}

% Existing methods  Existing methods mainly focus on domains via the lens of data sources. Recent works show that carefully curated data domains via prior knowledge or clustering are more effective for pre-training \citep{wettig2025organize,diao2025climb,rukhovich2025commute}. Under this scheme, existing works mainly develop static data mixtures via methods such as 

%% file: Sections/task_formulation.tex
\section{Domain Re-weighting as MDP}
In this section, we formally state domain re-weighting as a Markov Decision Process (MDP), defined as a tuple $(\mathcal{S}, \mathcal{A}, f, r, \rho_s, \rho_e)$. We describe each element as follows:

\paragraph{State Space}
The state space $\mathcal{S}$ is a continuous space consisting of all data distributions from previous domain reweighting steps. Specifically, for step $t$, the state $s_t\in \mathcal{S}$ has dimension $s_t\in \mathbb{R}^{t\times N}$, where $N$ is the dimension of the action space, determined by the definition of the domains.

\paragraph{Action Space}
The action space $\mathcal{A}$ is a continuous space denoting the data distribution in the current domain reweighting step. At step $t$, the action $a_t\in\mathcal{A}$ is a \textit{probability distribution} over the domain space: $a_t\in\mathbb{R}^N$ and $\sum_{i=1}^Na_t^i=1$, $a_t^i\geq0$.

\paragraph{Policy and Reward}
The policy function $f$ denotes Data Mixing Agent that determines the action at the current step. The feedback is modeled by the reward function $r$, determined by the target fields and the environment. With domain re-weighting modeled as an MDP, the policy function can be optimized via reinforcement learning.

\paragraph{Start and Terminate State}
The start state $\rho_s$ depends on the target model, reflecting the distribution of its pre-training data. The terminate state $\rho_e$ depends on manual setting or data scale, as the training process can often end with predefined token budgets or the exhaustion of target field data.

%% file: Sections/method.tex
\section{Data Mixing Agent}
\begin{figure*}[htbp]
\centering
\includegraphics[width=\textwidth]{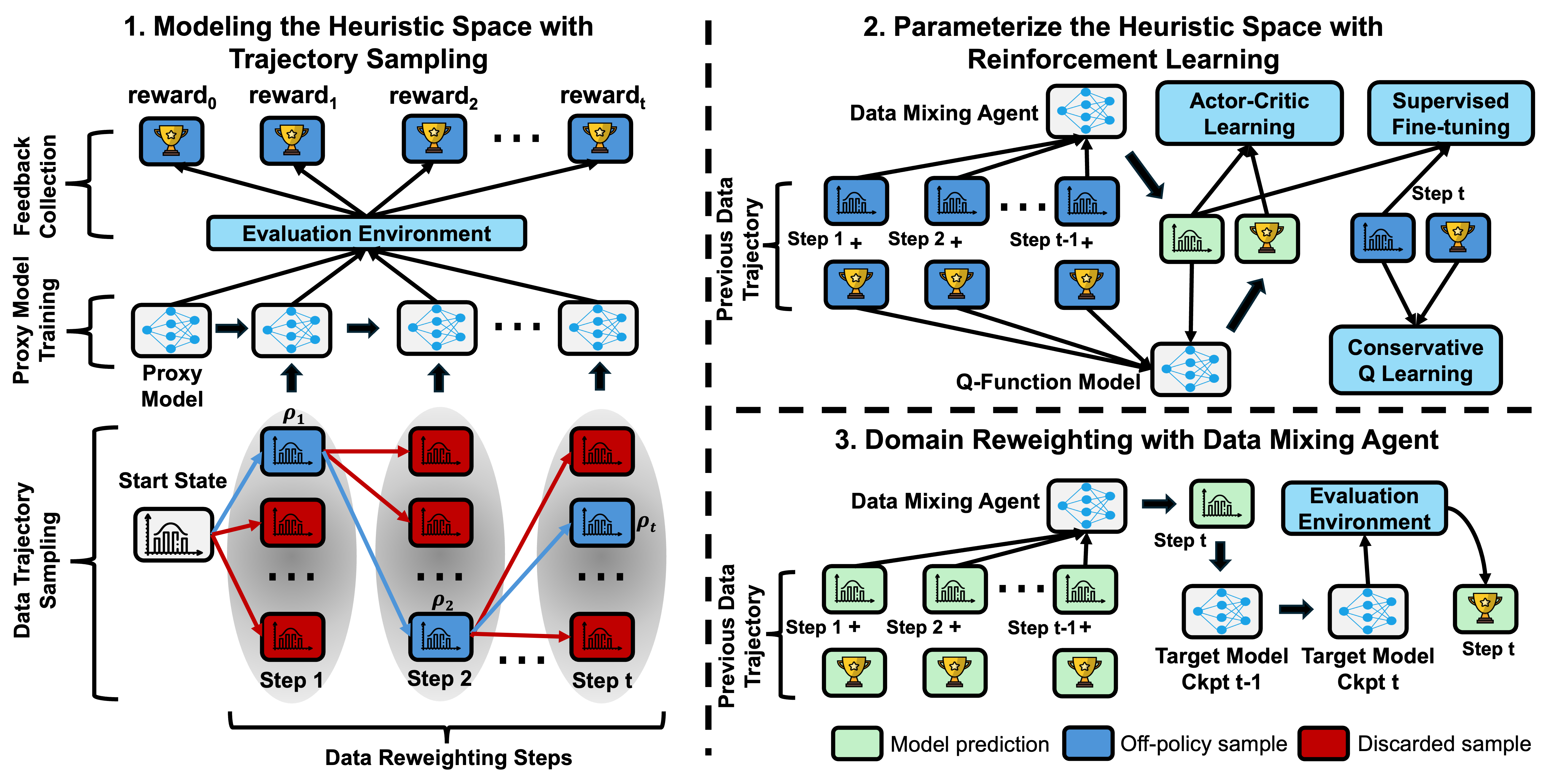}
\caption{An overview of Data Mixing Agent. We first sample data mixing trajectories and train small proxy models on them. Each model checkpoint obtains feedback from the evaluation environment. Secondly, the agent is optimized on these trajectories and feedback via CQL. When guiding continual pretraining, the agent determines the distribution for the next step on the fly.}
\label{fig:main_figure}
\end{figure*}

In this section, we introduce the methodology of Data Mixing Agent. An overview of the pipeline is shown in Fig. \ref{fig:main_figure}. \textbf{The full implementation details of all procedures are described in Appn. \ref{appn:implement_detail}.}

% \subsection{Modeling the Heuristic Space with Trajectory Sampling}\label{sec:data_collection}
\subsection{Trajectory Sampling}\label{sec:data_collection}
\paragraph{Action Space Definition}
We start by defining the action space $\mathcal{A}$, which is essential for trajectory sampling. While most methods define the space via data sources~\citep{xia2023sheared,luo2024velocitune,olmo20242}, recent work has emphasized the drawbacks of data overlap across domains~\citep{xi2025samplemix} and the unstructured nature~\citep{wettig2025organize} of source-based data clustering. Inspired by \citet{wettig2025organize}, we construct domains with the \href{https://huggingface.co/nvidia/domain-classifier}{Nvidia domain classifier}, which classifies the data from the source and target fields, each into 26 domains, leading to a 54-dimensional data distribution space. The definition of the 26 domains are shown in Fig. \ref{fig:vertical_histograms_example}.

\paragraph{Start State Estimation}
The start state $\rho_s$ can easily be determined when data from the source field are available. We randomly sample 1B tokens from the training data and utilize the same domain classifier to organize the data into the defined domains. The start state is estimated as a normalization over sample numbers in each domain. When data from the source field is unavailable~\citep{grattafiori2024llama,liu2024deepseek}, we explore using synthetic data from the target model to estimate the start state. We experiment on five Pythia models~\citep{biderman2023pythia} by estimating the distributions from their generated data and calculating their KL divergence with the ground-truth distribution. The results prove the effectiveness of using random samples from the target model as estimates for their start states. More details about the experiments and results are shown in Appn. \ref{appn:start_state_estimate}.

\paragraph{Data Mixing Trajectory Sampling}
We randomly sample data mixing trajectories as training data for modeling the heuristic space. The random sampling process is based on the following principle:
\begin{center}
\fcolorbox{black}{gray!10}{
\parbox{.9\linewidth}{
The data mixing trajectories should be well-distributed across the action space, ensuring coverage of actions that enhance and degrade model performance.
}
}
\end{center}
To ensure this principle, we design an inductive scoring algorithm to rate each sampled distribution, which is designed based on the following inductive biases: 1) The distribution at the current step should not deviate significantly from that of the previous step; 2) The distribution at each re-weighting step should align more closely with the target distribution; 3) The distribution at the current step should differ from sampled trajectories to encourage diversity; 4) The target distribution encourages reducing reliance on source-field data.
% \begin{itemize}
%     \item The distribution at the current step should not deviate significantly from that of the previous step;
%     \item The distribution at each domain re-weighting step should gradually align more closely with the target distribution;
%     \item The distribution at the current step should differ from previous sampled trajectories to encourage diversity;
%     \item The target distribution encourages trajectories to gradually reduce reliance on source-field data.
% \end{itemize}
During implementation, we use random tokens from the DCLM~\citep{li2024datacomp} as the source field data and the math split of the Dolmino-mix-1124~\citep{olmo20242} dataset as the target field data, obtaining 384 trajectories with various qualities. This algorithm is formally described in Appn. \ref{appn:trajectory_sampling}.

\paragraph{Environment Design and Feedback Collection}
The evaluation environment is curated to assess model checkpoints. It is lightweight yet accurately reflects target capabilities, providing effective supervision while minimizing computational overhead. Specifically, we select a small high-quality evaluation set $\mathcal{D}_i$ that well represents the $i$-th target field: $\{q_j, r_j\}_{j=1}^{|\mathcal{D}_i|}$. For the model checkpoint $\mathcal{M}$, we compute the average per-token log probability on all question-answer pairs to reflect model performance on the $i$-th target field:

\begin{small}
\begin{equation}\label{eqn:env_cal}
\mathcal{S}(\mathcal{M}, \mathcal{D}_i) = \frac{1}{|\mathcal{D}_i|} \sum_{(q_j, r_j) \in \mathcal{D}_i} \frac{1}{|r_j|}\log P_{\mathcal{M}}(r_j \mid q_j)
\end{equation}
\end{small}

The final environment feedback returns a vector-style assessment for model $\mathcal{M}$:

\begin{small}
\begin{equation}\label{eqn:env_cal_1}
\mathcal{R}(\mathcal{M}) = \left[\mathcal{S}(\mathcal{M}, \mathcal{D}_1), \mathcal{S}(\mathcal{M}, \mathcal{D}_2),...,\mathcal{S}(\mathcal{M}, \mathcal{D}_{|\mathcal{D}|})\right]
\end{equation}
\end{small}

During implementation, the environment assesses the general capability via the validation set of the MMLU~\citep{hendrycks2020measuring} dataset and the math reasoning capability via a subset of the MATH~\citep{hendrycks2021measuring} dataset. Leveraging this evaluation environment, we collect feedback data by training a small proxy model $\mathcal{M}_p$ on each sampled trajectory from scratch. The model checkpoint is evaluated on this environment at each data reweighting step. For the $i$-th data mixing distribution $\rho_i\in\tau$, we obtain a tuple $(\rho_i, \mathcal{R}(\mathcal{M}_p^i))$, where $\mathcal{M}_p^i$ denotes the model checkpoint after training on the $i$-th step.

\subsection{Modeling the Heuristic Space with RL}
We expect the sampled data mixing trajectories and the feedback to well represent the heuristic space for domain reweighting. We parameterize these heuristics by training an agent model on these trajectories in a reinforcement learning-based paradigm.

\paragraph{Agent Model Structure}
We determine the model structure for the data mixing agent with the following principles:
1) effectively model temporal sequences and support long-range interactions between distributions; 2) be lightweight to prevent unacceptable latency and computation overhead.
% \begin{itemize}
%     \item It should effectively model temporal sequences and support long-range interactions between distributions;
    
%     \item It should be lightweight to prevent unacceptable latency and computation overhead during target model training.
% \end{itemize}
We utilize the Transformer~\citep{vaswani2017attention} decoder architecture, which is widely used in time series forecasting~\citep{zhang2024large,li2025mira} and facilitates long-range interactions between data points with dot-product attention. We stack two Transformer layers followed by a linear layer and Softmax to project the representations into the action space, with merely 2.1M parameters. Formally, at data reweighting step $t$, the agent $f$ predicts the domain distribution with the previous trajectory and environment feedback as follows:

\begin{small}
\begin{equation}\label{eqn:agent_predict}
\begin{aligned}
&\rho_t = f(\tilde{\rho}_0,\tilde{\rho}_1,...,\tilde{\rho}_{t-1})\\
&\tilde{{\rho}}_i=\left[\rho_i;\mathcal{R}(\mathcal{M}_i)\right], i=0,...,t-1
\end{aligned}
\end{equation}
\end{small}

where $;$ denotes concatenation, $\tilde{\rho}_i\in\mathbb{R}^{N+|\mathcal{D}|}$ denotes the input feature in the data reweighting step $i$, and $\rho_t$ denotes the agent's output action at step $t$.

\paragraph{Off-policy Optimization with Conservative Q-Learning}
We first perform a warm-up on the randomly initialized agent model with Supervised Fine-Tuning (SFT) to reduce the later parameter searching space. Details about the SFT process is shown in Appn. \ref{appn:sft}.
Based on the warmed-up model, we parameterize the heuristic space via reinforcement learning, where the algorithm is selected based on the following two principles:
1) The training process is offline and off-policy;
2) The agent's actions are sampled from a continuous domain space.
% \begin{itemize}
%     \item The training process is offline and off-policy, with data collected from other proxy models and no access to the evaluation environment;

%     \item The agent's actions are probability distributions sampled from a continuous domain space.
% \end{itemize}
We select Conservative Q-Learning (CQL)~\citep{kumar2020conservative} as the optimization algorithm. CQL prevents overestimation of Q-values for out-of-distribution actions by introducing a conservative penalty for the Q-function optimization process.
The agent model is trained in an actor-critic~\citep{sutton1999policy} structure until convergence, where the agent acts as the actor model, and the critic model (Q-function) is initialized from scratch with another neural network. Details about the CQL training process are shown in Appn. \ref{appn:cql}.

\subsection{Domain Reweighting with Data Mixing Agent}
The system pipeline of continual pretraining with Data Mixing Agent is described in Algorithm \ref{algo:data_mixing_agent}.
The agent directly determines the distribution for the next domain re-weighting step on the fly, considering the previous states in the data mixing trajectory and the corresponding environment feedback. This MDP continues until the target data is fully leveraged or a predetermined computation budget is reached. We expect the agent to optimally curate the training recipe by balancing performance across all target fields while minimizing the use of source-field data tokens to reduce computational cost. 
\textbf{We also expect the agent’s learned heuristics to generalize to unseen target models, data mixtures, and even target domains.} This generalization is crucial to avoid repeated trajectory sampling and agent retraining when adapting to new continual pre-training scenarios, thereby significantly reducing overall computational cost.

%% file: Sections/experiments.tex
\input{Tables/main-results}

\section{Experiments}
\subsection{Experimental Settings}\label{sec:experiment_set}
\textbf{Full justifications for implementation and experimental settings are described in Appn. \ref{appn:experiment_setting}.}

\paragraph{Target Models}
Since most open-source LLMs have already been optimized on math reasoning or code generation, we pre-train 3 models from scratch, with the same LLaMA model architecture~\citep{grattafiori2024llama} with 3B parameters, on 100B random tokens from the DCLM~\citep{li2024datacomp}, Fineweb-Edu~\citep{penedo2024fineweb}, and Nemotron-CC~\citep{su2024nemotron} dataset, resulting in 3 target models: \textbf{LLaMA-DCLM}, \textbf{LLaMA-FWE}, and \textbf{LLaMA-Nemo}. We also include the \textbf{Pythia-1.4B} model~\citep{biderman2023pythia} to evaluate performance on existing open-source models.

\paragraph{Baseline Methods}
We compare Data Mixing Agent (\textbf{DataAgent$_{RL}$}) with the following baseline methods: \textbf{Base Model}: direct evaluation of the target models; \textbf{Naive Training}: continual training solely on target field data; \textbf{RegMix}~\citep{liu2024regmix}: trains small proxy models on multiple distributions and fit a regression model to determine the optimal recipe; \textbf{Dynamic Batch Loading (DBL)}~\citep{xia2023sheared}: dynamic domain re-weighting based on excess losses across different domains; \textbf{DataAgent$_{SFT}$}: the data mixing agent model without the reinforcement learning process. DBL can only be applied to the data source-based experiments since it requires an evaluation set for each domain. 

\paragraph{Target Model Training Data}
For source field data, the pretrained LLaMA models utilize their corresponding pre-training data, each with 100B tokens. We use synthetic data from the Pythia-1.4B with quality filters as the source field data for itself. For math reasoning, we select the math split (10B tokens) of the Dolmino-mix-1124~\citep{olmo20242}. For code generation, we select the GitHub training split of SlimPajama-DC~\citep{shen2023slimpajama} with 30B tokens.

\paragraph{Evaluation Benchmarks}
We evaluate general capabilities on the MMLU~\citep{hendrycks2020measuring}, HellaSwag~\citep{zellers2019hellaswag} (Hella.), OpenBookQA~\citep{mihaylov2018can} (OBQA), Winogrande~\citep{sakaguchi2021winogrande} (Wino.), ARC-Challenge~\citep{clark2018think} (ARC-C), PiQA~\citep{bisk2020piqa}, SciQ~\citep{welbl2017crowdsourcing}, and LogiQA~\citep{liu2020logiqa} benchmarks. We evaluate math reasoning capabilities on the GSM8K~\citep{cobbe2021training}, MATH~\citep{hendrycks2021measuring}, Minerva~\citep{lewkowycz2022solving}, and MathQA~\citep{amini2019mathqa} benchmarks. We evaluate code generation capabilities on the HumanEval~\citep{chen2021evaluating} and MBPP~\citep{austin2021program} benchmarks.

\input{Tables/code-results}

\paragraph{Target Model Training Setting}
The agent is trained on a 52-dimensional domain reweighting space. To evaluate in different domain spaces, we further employ the agent on the data source-based action space (source and target). We use the GitHub validation split of the SlimPajama-DC dataset when building the environment for code generation.

\subsection{Evaluation Results on Math Reasoning}
Evaluation results on the math reasoning target field are shown in Table \ref{tab:main-results}. We have the following observations:

\textbf{Data Mixing Agent significantly outperforms other methods in balanced performance across fields.} In Table \ref{tab:2D-main-results}, for the LLaMA-DCLM target model, DataAgent$_{RL}$ achieves the best average performance 54.04\% and 33.02\% on general/math benchmarks, outperforming the base model in general ability and the naively trained model on math reasoning. Overall, DataAgent$_{RL}$ achieves 47.03\% on average, surpassing RegMix by 3.02\%, DBL by 3.53\%, and the base model by 8.88\%. DataAgent$_{RL}$ also outperforms DataAgent$_{SFT}$ by a large margin of 2.08\%, proving reinforcement learning with CQL as a crucial step for effective heuristics compared to the inductive biases in the SFT stage. All domain reweighting methods significantly reduce performance variance by over 200 compared to the base model. Data mixing agent further achieves the best results in variance, further demonstrating its advantage in stable performance across fields.

\textbf{The capabilities of data mixing agents can generalize across target models, source-field data, and domain spaces without retraining.} Though the agent is trained on the 52-dimensional data reweighting space with trajectories sampled with the DCLM data, it effectively guides domain reweighting for three target models across 2 domain definitions. On the 2-dimensional domain space (Table \ref{tab:2D-main-results}), DataAgent$_{RL}$ outperforms DBL by an average of 2.37\% on unseen target models LLaMA-FWE and LLaMA-Nemo. On the 52-dimensional space (Table \ref{tab:nvidia-main-results}), DataAgent$_{RL}$ outperforms RegMix by an average of 1.41\%. These results reflect data- and model-agnostic heuristics, enabling the agent to generalize well without re-training.

Full results and analysis on the Pythia-1.4B model and synthetic source field data are presented in Appn. \ref{appn:result_math}.

\subsection{Evaluation Results on Code Generation}
We directly utilize the agent trained on math reasoning to guide domain reweighting for code generation. The results are shown in Table \ref{tab:code_result}. We have the following observations:

\textbf{Data Mixing Agent can generalize across target fields without retraining.} DataAgent$_{RL}$ achieves the best average performance of 46.3\%, significantly outperforming RegMix by 1.45\% and DBL by 2.78\%, which are directly optimized on code generation. It shows that heuristics can be directly transferred without modifying the weights of the agent. In code generation, DataAgent$_{RL}$ outperforms naive training by 6.22\%, while in Table \ref{tab:2D-main-results-appn}, the advantage is 8.52\%, mainly due to the fact that applying DataAgent$_{RL}$ to code generation leads to worse performance on general benchmarks compared to math reasoning, which indicates the existence of heuristics that are dependent on the target field and the potential misalignment when converting them to a new target field.

Full results and analysis on the Pythia-1.4B model and synthetic source field data are presented in Appn. \ref{appn:result_code}.

\begin{figure*}[htpb]
\centering
\includegraphics[width=15cm,height=9cm]{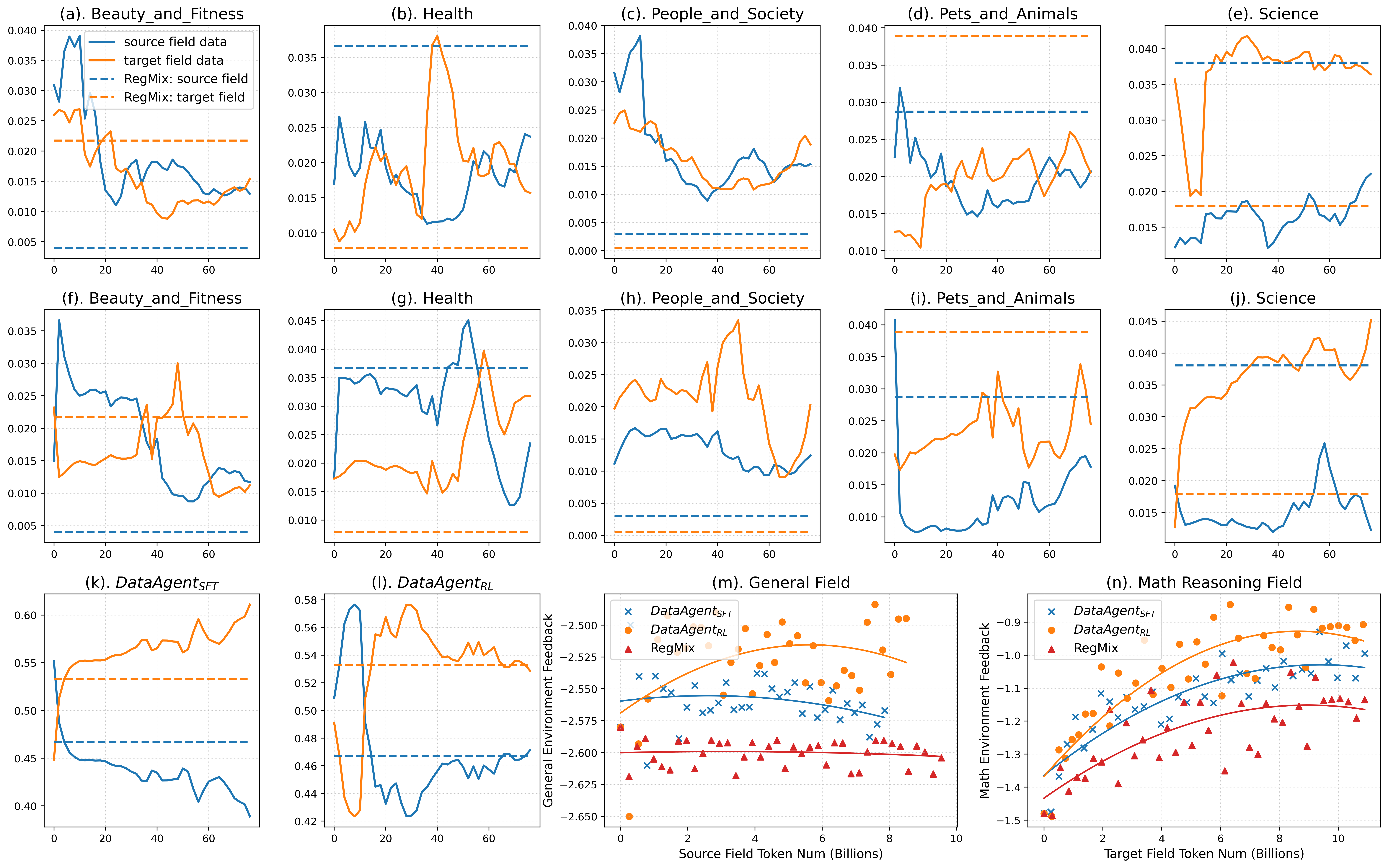}
\caption{Training on LLaMA-DCLM and math reasoning, (a)-(e) presents DataAgent$_{RL}$'s output trajectories on 5 domains within the 52-dimensional space. (f)-(j) presents DataAgent$_{SFT}$'s trajectories. (k)-(l) presents the two agents' trajectories on the 2-dimensional domain space. (a)-(l) share the legend within (a). Dashed lines show RegMix outputs. (m)-(n) present the dynamics of model performance on the evaluation environment with increasing training data in the corresponding field.}
\label{fig:all_in_one}
\end{figure*}

\subsection{Comparisons on Training Efficiency}
In Table \ref{tab:main-results} and \ref{tab:code_result}, we provide the training time of the main experiments measured in GPU hours to compare their efficiency.

\textbf{Data Mixing Agent requires fewer GPU hours for training than baseline methods, demonstrating higher efficiency despite the addition of agent models.} For instance, DataAgent$_{RL}$ saves an average of 106.33 GPU hours compared to RegMix in math reasoning, and 264.29 GPU hours in code generation. This efficiency primarily stems from two factors: (1) the agent model is extremely lightweight, adding negligible computational overhead and latency during inference; and (2) it is trained to reduce dependence on source-field data, thereby lowering the overall token budget by incorporating less source-field data. Further discussion of these savings is provided in Sec.~\ref{sec:data_efficiency}.

Training agent models involves additional computational cost from trajectory sampling, where training proxy models on 384 trajectories and their evaluation require 1996.08 GPU hours in total. In contrast, training the lightweight agent model with SFT and CQL finishes within 10 minutes. \textbf{Importantly, the agent training process is needed only once, and our experiments demonstrate that the Data Mixing Agent generalizes across source and target fields, target models, and domain spaces without re-training.}

\subsection{Analysis on Reweighting Trajectories}
We show the data mixing trajectories guided by the agent to provide more intuition on its learned heuristics. The 2-dimensional results and part of 52-dimensional results are shown in Fig. \ref{fig:all_in_one}(a)-(l). The full trajectories of all domains and a more detailed analysis are presented in Appn. \ref{appn:result_traj_analysis}.

\textbf{Data Mixing Agent follows a less-to-more trend for target field data along, but DataAgent$_{RL}$ adopts a more fine-grained approach.} In Fig. \ref{fig:all_in_one}(k)-(l), both the DataAgent$_{RL}$ and DataAgent$_{SFT}$ models increase target field data, where DataAgent$_{SFT}$ shows a radical increasing trend for math data, almost monotonically from 45\% to over 60\%. DataAgent$_{RL}$ adopts a fine-grained three-stage strategy: (1) \textbf{Early warm-up} to prioritize source field data; (2) \textbf{Mid training} to rapidly increase target field data; (3) \textbf{Final stage} to gradually reintroduce more source field data, stabilizing around the optimal RegMix weights. The superior performance of DataAgent$_{RL}$ proves the effectiveness of optimization on various trajectories via reinforcement learning. In Fig. \ref{fig:all_in_one}(a)-(j), 
the 52-dimensional results further strengthen the above arguments. In all domains, DataAgent$_{SFT}$ organizes the target field data from 60\% of the trajectories to increase monotonically, while DataAgent$_{RL}$ introduces more subtle strategies in 80\% of the trajectories.

\textbf{Data Mixing Agent learns heuristics that correspond to human intuition.} \citet{wettig2025organize} summarized the top-3 domains that benefit the MMLU performance: $Science\&Tech.$, $Health$, and $Politics$. In Fig. \ref{fig:all_in_one}(a)-(e), we observe a significant uplift of data in the corresponding domains compared to RegMix distributions: $Science$, $Health$, and $People\&Society$.  \citet{wettig2025organize} also enumerated domains that hurt performance, such as $Fashion\&Beauty$. where DataAgent$_{RL}$ also conveys an explicit downsampling process. These observations further encourage the discovery of new general heuristics. For example, DataAgent$_{RL}$ continuously reduces data from both source and target fields in the $Pets\&Animals$, indicating its less importance in general or math reasoning capabilities.

\subsection{Data Efficiency}\label{sec:data_efficiency}
We explore how efficiently the agents leverage the source and target field data to improve or preserve model capabilities in the corresponding fields. We record the performance dynamics of LLaMA-DCLM with increasing training data. The results are shown in Fig. \ref{fig:all_in_one}(m)-(n). A more detailed analysis is presented in Appn. \ref{appn:result_efficiency}.

\textbf{Data Mixing Agent leverages general field data more efficiently than RegMix, better preserving model capabilities in the source field.} As shown in Fig. \ref{fig:all_in_one}(m), the agent obtains higher general feedback values from the environment at most token budgets for the source field. DataAgent$_{RL}$ further outperforms DataAgent$_{SFT}$, with feedback values fluctuating around -2.525. Notably, DataAgent$_{RL}$ shows higher variance along the domain reweighting trajectory, reflecting its active strategies in adjusting domain reweighting distributions to improve source field capabilities. Its superior performance on general benchmarks (Table \ref{tab:2D-main-results}) indicates the effectiveness of such strategies.

\textbf{Data Mixing Agent leverages data from the math reasoning field more efficiently than RegMix, achieving higher performance in the target field.} As shown in Fig. \ref{fig:all_in_one}(n), though all methods show logarithmic-scale improvements on math reasoning feedback, the agent methods show a faster momentum in increasing feedback values. RegMix performance stabilizes around -1.2 while both data mixing agent methods achieve performance over -1.1. DataAgent$_{RL}$ further outperforms DataAgent$_{SFT}$ with the optimized feedback values over -1.0. Overall, DataAgent$_{RL}$ can coordinate the source and target field data to improve performance on multiple target capabilities.

\textbf{Data Mixing Agent achieves balanced continual pre-training performance with less reliance on data from the source field.}. While we set a total training budget of 21B tokens, DataAgent$_{RL}$ triggers an early stopping at 19.92B tokens, and DataAgent$_{SFT}$ triggers an early stopping at 18.86B tokens, due to the exhaustion of the target field data. These results show that the agent can achieve superior performance than RegMix in both the general and math reasoning fields while saving up to a training token budget of 2.14B.

%% file: Tables/main-results.tex
\begin{table*}[!hbt]
\centering

% ---------- 2D dimension results ----------
\begin{subtable}{\textwidth}
\centering
\resizebox{1.\textwidth}{!}{
\begin{tabular}{lccc|ccccccccc|ccccc}
\toprule
\multirow{2}{*}{\textbf{Method}} & \multirow{2}{*}{\textbf{GPU Hrs}} & \multirow{2}{*}{\textbf{Avg.}$\uparrow$} & \multirow{2}{*}{\textbf{Var.}$\downarrow$} & \multicolumn{9}{c}{\textbf{General Benchmarks}} & \multicolumn{5}{c}{\textbf{Math Benchmarks}} \\
& & & & MMLU & Hella. & OBQA & Wino. & ARC-C & PiQA & SciQ & LogiQA & Avg. & GSM8K & Minerva & MATH & MathQA & Avg. \\
\midrule
\multicolumn{17}{c}{\textbf{LLaMA-DCLM}}\\
Base Model & -- & 38.15 & 780.62 & 34.5 & \textbf{64.5} & 37.0 & 61.56 & 36.69 & \textbf{75.84} & 84.2 & 28.11 & 52.8 & 2.55 & 4.1 & 4.22 & 24.52 & 8.85 \\
Naive Training & 874.88 & 38.51 & -- & 27.11 & 37.0 & 28.2 & 54.22 & 28.58 & 60.28 & 68.7 & 26.88 & 41.37 & 59.21 & \textbf{16.16} & \textbf{22.85} & 32.96 & 32.80 \\
RegMix & 1987.84 & 44.01 & 559.83 & 30.42 & 59.72 & 36.6 & 61.72 & 34.73 & 73.88 & 85.1 & 28.73 & 51.36 & 55.87 & 11.5 & 17.7 & 32.16 & 29.31\\
DBL & 1906.8 & 43.50 & 575.91 & 29.0 & 55.66 & 35.0 & \textbf{63.64} & 32.11 & 72.5 & \textbf{88.42} & 28.89 & 50.65 & 56.22 & 12.2 & 18.24 & 30.14 & 29.2\\
DataAgent$_{SFT}$ & 1880.32 & 44.95 & 563.27 & 33.81 & 60.23 & 34.2 & 60.43 & 36.26 & 73.28 & 87.3 & 29.33 & 51.86 & 57.84 & 12.7 & 21.3 & 32.73 & 31.14\\
DataAgent$_{RL}$ & 1891.84 & \textbf{47.03} & \textbf{547.66} & \textbf{34.06} & 63.38 & \textbf{42.14} & 62.35 & \textbf{36.92} & 74.85 & 87.89 & \textbf{30.29} & \textbf{54.04} & \textbf{59.24} & 14.8 & 22.75 & \textbf{35.3} & \textbf{33.02}\\
\midrule
\multicolumn{17}{c}{\textbf{LLaMA-FWE}}\\
Base Model & -- & 37.65 & 781.34 & 34.47 & 60.52 & 37.8 & 57.77 & 40.53 & 74.21 & 85.4 & \textbf{28.26} & 52.37 & 2.5 & 2.52 & 4.06 & 23.72 & 8.2 \\
Naive Training & 852.1 & 38.51 & -- & 27.25 & 37.03 & 28.4 & 53.51 & 26.96 & 61.1 & 69.6 & 28.11 & 41.5 & \textbf{58.91} & 12.58 & \textbf{24.7} & \textbf{33.94} & \textbf{32.53}\\
RegMix & 1948.26 & 43.83 & 557.18 & 32.83 & 60.2 & 36.45 & 54.51 & 37.17 & 71.72 & 86.5 & 28.0 & 50.92 & 55.9 & 12.2 & 20.35 & 30.1 & 29.64 \\
DBL & 1883.58 & 43.92 & 566.47 & 31.27 & 56.58 & \textbf{40.7} & 56.27 & 38.32 & 71.18 & \textbf{88.26} & 26.0 & 51.07 & 54.2 & 12.05 & 20.72 & 31.52 & 29.62\\
DataAgent$_{SFT}$ & 1887.82 & 45.23 & 552.61 & \textbf{34.65} & \textbf{60.83} & 38.28 & 59.3 & \textbf{40.8} & \textbf{74.6} & 85.6 & 26.96 & \textbf{52.63} & 56.26 & 12.19 & 21.92 & 31.32 & 30.42\\
DataAgent$_{RL}$ & 1838.2 & \textbf{45.48} & \textbf{540.83} & 33.78 & 60.44 & 38.8 & \textbf{59.59} & 38.89 & 73.12 & 84.9 & 27.49 & 52.13 & 58.07 & \textbf{13.46} & 23.96 & 33.28 & 32.19\\
\midrule
\multicolumn{17}{c}{\textbf{LLaMA-Nemo}}\\
Base Model & -- & 38.22 & 782.95 & 34.22 & 64.51 & 37.6 & 59.12 & 36.26 & \textbf{75.57} & \textbf{88.4} & 26.73 & 52.8 & 2.5 & 4.85 & 5.3 & 23.62 & 9.07\\
Naive Training & 870.4 & 37.86 & -- & 27.06 & 37.4 & 28.0 & 52.88 & 27.05 & 59.74 & 68.6 & 25.19 & 40.74 & \textbf{59.05} & \textbf{12.3} & \textbf{24.52} & \textbf{32.53} & \textbf{32.1}\\
RegMix & 1925.23 & 44.13 & 546.24 & \textbf{35.03} & \textbf{65.15} & 35.9 & 59.83 & 36.45 & 72.85 & 86.2 & 28.88 & 52.54 & 49.39 & 10.7 & 19.23 & 30.0 & 27.33\\
DBL & 1920.21 & 42.63 & 522.88 & 33.2 & 64.82 & 34.0 & \textbf{62.06} & \textbf{39.23} & 70.79 & 78.11 & 24.0 & 50.77 & 45.91 & 10.74 & 20.01 & 28.74 & 26.35\\
DataAgent$_{SFT}$ & 1887.25 & 44.12 & 538.44 & 34.06 & 64.25 & 39.04 & 60.4 & 38.1 & 74.17 & 86.75 & 29.16 & 53.24 & 47.81 & 9.07 & 17.8 & 28.86 & 25.89\\
DataAgent$_{RL}$ & 1812.31 & \textbf{45.8} & \textbf{520.71} & 34.27 & 63.95 & \textbf{39.8} & 61.58 & 38.74 & 74.49 & 86.9 & \textbf{29.95} & \textbf{53.71} & 54.28 & 10.83 & 22.94 & 31.85 & 29.98\\
\bottomrule
\end{tabular}}
\caption{Model performances on the 2-dimensional data reweighting space based on data sources.}\label{tab:2D-main-results}
\end{subtable}

\vspace{0.8em}

% ---------- Nvidia results ----------
\begin{subtable}{\textwidth}
\centering
\resizebox{1.\textwidth}{!}{
\begin{tabular}{lccc|ccccccccc|ccccc}
\toprule
\multirow{2}{*}{\textbf{Method}} & \multirow{2}{*}{\textbf{GPU Hrs}} & \multirow{2}{*}{\textbf{Avg.}$\uparrow$} & \multirow{2}{*}{\textbf{Var.}$\downarrow$} & \multicolumn{9}{c}{\textbf{General Benchmarks}} & \multicolumn{5}{c}{\textbf{Math Benchmarks}} \\
& & & & MMLU & Hella. & OBQA & Wino. & ARC-C & PiQA & SciQ & LogiQA & Avg. & GSM8K & Minerva & MATH & MathQA & Avg. \\
\midrule
\multicolumn{17}{c}{\textbf{LLaMA-DCLM}}\\
Base Model & -- & 38.15 & 780.62 & \textbf{34.5} & \textbf{64.5} & 37.0 & 61.56 & 36.69 & \textbf{75.84} & 84.2 & 28.11 & 52.8 & 2.55 & 4.1 & 4.22 & 24.52 & 8.85 \\
Naive Training & 874.88 & 38.51 & -- & 27.11 & 37.0 & 28.2 & 54.22 & 28.58 & 60.28 & 68.7 & 26.88 & 41.37 & \textbf{59.21} & 16.16 & 22.85 & \textbf{32.96} & \textbf{32.80} \\
RegMix & 2306.14 & 44.67 & 571.39 & 34.38 & 62.17 & 38.2 & 61.93 & \textbf{36.95} & 74.97 & 87.5 & 29.49 & 53.19 & 55.78 & 10.36 & 15.75 & 28.67 & 27.64 \\
DataAgent$_{SFT}$ & 2174.26 & 45.75 & 560.74 & 34.47 & 63.36 & 40.6 & 62.35 & 35.87 & 74.32 & \textbf{89.2} & \textbf{29.96} & 53.77 & 56.77 & 11.12 & 18.76 & 32.3 & 29.74\\
DataAgent$_{RL}$ & 2202.88 & \textbf{46.84} & \textbf{552.11} & 32.99 & 62.64 & \textbf{41.6} & \textbf{63.98} & 36.64 & 73.5 & 89.1 & 31.5 & \textbf{53.99} & 59.04 & \textbf{16.48} & \textbf{22.9} & 31.72 & 32.54 \\
\midrule
\multicolumn{17}{c}{\textbf{LLaMA-FWE}}\\
Base Model & -- & 37.65 & 781.34 & 34.47 & 60.52 & 37.8 & 57.77 & \textbf{40.53} & 74.21 & 85.4 & 28.26 & \textbf{52.37} & 2.5 & 2.52 & 4.06 & 23.72 & 8.2 \\
Naive Training & 852.1 & 38.51 & -- & 27.25 & 37.03 & 28.4 & 53.51 & 26.96 & 61.1 & 69.6 & 28.11 & 41.5 & 58.91 & 12.58 & \textbf{24.7} & \textbf{33.94} & 32.53\\
RegMix & 2276.97 & 43.78 & 553.76 & \textbf{35.64} & 57.64 & \textbf{39.4} & 57.56 & 38.2 & 67.4 & 85.03 & 29.34 & 51.28 & 53.68 & 10.84 & 20.35 & 30.0 & 28.72\\
DataAgent$_{SFT}$ & 2102.56 & 45.21 & 541.08 & 34.27 & 61.2 & 37.95 & 58.57 & 40.28 & \textbf{75.17} & 85.8 & 28.23 & 52.68 & 54.5 & 13.27 & 22.17 & 31.1 & 30.26\\
DataAgent$_{RL}$ & 2162.57 & \textbf{45.55} & \textbf{532.47} & 32.55 & 58.64 & 35.8 & \textbf{59.42} & 39.76 & 73.34 & \textbf{87.2} & \textbf{29.35} & 52.01 & \textbf{58.96} & \textbf{14.68} & 23.62 & 33.32 & \textbf{32.65}\\
\midrule
\multicolumn{17}{c}{\textbf{LLaMA-Nemo}}\\
Base Model & -- & 38.22 & 782.95 & 34.22 & 64.51 & 37.6 & 59.12 & 36.26 & \textbf{75.57} & 88.4 & 26.73 & 52.8 & 2.5 & 4.85 & 5.3 & 23.62 & 9.07\\
Naive Training & 870.4 & 37.86 & -- & 27.06 & 37.4 & 28.0 & 52.88 & 27.05 & 59.74 & 68.6 & 25.19 & 40.74 & \textbf{59.05} & \textbf{12.3} & \textbf{24.52} & \textbf{32.53} & \textbf{32.1}\\
RegMix & 2276.17 & 44.86 & 559.63 & \textbf{35.68} & \textbf{66.4} & 41.4 & \textbf{61.56} & 35.82 & 72.4 & \textbf{87.53} & 29.34 & \textbf{53.77} & 48.17 & 10.91 & 19.03 & 30.04 & 27.04\\
DataAgent$_{SFT}$ & 2144.15 & 45.16 & 544.92 & 34.8 & 64.01 & 38.2 & 60.73 & 36.9 & 73.92 & 87.3 & \textbf{29.37} & 53.15 & 52.5 & 10.7 & 20.77 & 32.71 & 29.17\\
DataAgent$_{RL}$ & 2160.44 & \textbf{45.9} & \textbf{536.18} & 33.89 & 62.7 & \textbf{42.6} & 59.82 & \textbf{40.0} & 74.86 & 84.34 & 28.29 & 53.31 & 56.78 & 11.82 & 23.61 & 32.03 & 31.06\\
\bottomrule
\end{tabular}}
\caption{Model performances on the 52-dimensional data reweighting space based on the Nvidia domain classifier.}\label{tab:nvidia-main-results}
\end{subtable}
\caption{Results of math reasoning on 12 benchmarks. We apply DBL only on the 2D domain space, due to the lack of evaluation sets for all domains in the 52D space. "GPU Hrs" denotes training time measured in GPU hours. "Avg." and "Var." denote the average and variance of each method on all benchmarks.}\label{tab:main-results}
\end{table*}

%% file: Tables/code-results.tex
\begin{table*}[!hbt]
\centering
\resizebox{1.\textwidth}{!}{
\begin{tabular}{lcc|ccccccccc|ccc}
\toprule
\multirow{2}{*}{\textbf{Method}} & \multirow{2}{*}{\textbf{GPU Hrs}} & \multirow{2}{*}{\textbf{Avg.}} & \multicolumn{9}{c}{\textbf{General Benchmarks}} & \multicolumn{3}{c}{\textbf{Code Benchmarks}} \\
& & & MMLU & Hella. & OBQA & Wino. & ARC-C & PiQA & SciQ & LogiQA & Avg. & HumanEval & MBPP & Avg. \\
\midrule
Base Model & -- & 44.52 & \textbf{34.5} & \textbf{64.5} & 37.0 & \textbf{61.56} & \textbf{36.69} & 75.84 & \textbf{84.2} & 28.11 & \textbf{52.8} & 8.6 & 14.2 & 11.4 \\
Naive Training & 3390.77 & 40.08 & 27.6 & 42.96 & 29.37 & 53.1 & 24.76 & 70.5 & 62.95 & 24.46 & 41.96 & \textbf{27.3} & \textbf{37.8} & \textbf{32.55} \\
RegMix & 5726.11 & 44.85 & 31.22 & 57.13 & 33.4 & 57.43 & 29.05 & \textbf{76.1} & 82.09 & 29.33 & 49.47 & 21.1 & 31.6 & 26.35 \\
DBL & 5480.74 & 43.52 & 31.7 & 55.7 & 35.2 & 53.8 & 31.28 & 68.58 & 78.26 & \textbf{29.8} & 48.04 & 20.6 & 30.3 & 25.45\\
DataAgent$_{SFT}$ & 5318.37 & 45.07 & 32.69 & 63.43 & 35.0 & 49.88 & 33.46 & 72.85 & 78.66 & 29.61 & 49.45 & 22.4 & 32.8 & 27.6 \\
DataAgent$_{RL}$ & 5461.82 & \textbf{46.3} & 33.84 & 63.79 & \textbf{37.8} & 57.3 & 35.05 & 73.2 & 78.57 & 27.34 & 50.86 & 22.0 & 34.1 & 28.05 \\
\bottomrule
\end{tabular}}
\caption{Evaluation results on the code generation target field, reflected on 10 benchmarks. The data is reweighted on the 2-dimensional data source-based domain space and the LLaMA-DCLM target model.}
\label{tab:code_result}
\end{table*}

%% file: Sections/related_work_simple.tex
\section{Related Work}
Continual pre-training has been widely applied in domain-specific LLM adaptation~\citep{shao2024deepseekmath,guo2024deepseek,hui2024qwen2,xie2024finben,lin2025sigma,tu2024towards}, often facing the catastrophic forgetting problem~\citep{hui2024qwen2,lin2025sigma,luo2023empirical,yang2024qwen2}. To mitigate this, data mixing strategies have merged~\citep{xie2023doremi,liu2024regmix,xia2023sheared,luo2024velocitune} to balance model performance across fields. Recent studies also highlight the role of domain space definitions in improving re-weighting performance~\citep{wettig2025organize,rukhovich2025commute,diao2025climb,xi2025samplemix}. More details in Appn. \ref{appn:related_work}.

%% file: Sections/conclusion.tex
\section{Conclusion}
We propose Data Mixing Agent, the first model-based domain reweighting method for continual pre-training. By learning general heuristics through reinforcement learning on sampled mixing trajectories with evaluation feedback, it consistently outperforms strong baselines on 12 general and math reasoning benchmarks. The learned strategies generalize across source data, models, domain spaces, and new fields without retraining, while aligning with human intuitions and achieving superior performance with less data.

%% file: Sections/limitations.tex
\section*{Limitations}
The limitations of this work are threefold. Firstly, due to the high cost of continual pre-training and limited computational resources, we evaluated Data Mixing Agents on only two domain reweighting spaces, four target models, and two target fields (math reasoning and code generation). Future work will extend the evaluation to a broader range of settings, particularly to test generalization across more target fields and domain spaces. Secondly, our evaluation focused exclusively on continual pre-training scenarios. An important direction for future work is to explore the applicability of Data Mixing Agents in large-scale pre-training. Thirdly, although Data Mixing Agents generalize well across related fields, they may underperform when the target field exhibits a significant distribution shift from the source field (e.g., from math reasoning to the medical domain). We leave a detailed analysis of such cases, along with potential optimizations, to future work.

%% file: Appendix/sections/heuristic.tex
\section{Agent Model}
\label{appn:general_heuristic}
To prove that data mixing agent is qualified as an agent model, we’d like to start with Google’s definition of the term AI agent:

\begin{center}
\fcolorbox{black}{gray!10}{
\parbox{.9\linewidth}{
AI agents are software systems that use AI to pursue goals and complete tasks on behalf of users. They show reasoning, planning, and memory and have a level of autonomy to make decisions, learn, and adapt.
}
}
\end{center}

Based on this definition, we prove data mixing agent can be qualified as an agent for the following reasons:

\textbf{Software systems}: Data mixing agent is a Transformer-based machine learning system.
Goals and Tasks: Data mixing agent takes the previous training history as inputs and output the optimal data mixing recipe for the next domain reweighting step.

\textbf{Reasoning, planning, and memory}: Data mixing agent memorizes previous data mixing trajectories and reasons over them to plan for the future training recipe.

\textbf{Learn, and adapt}: Our experiments prove that data mixing agent can generalize well across fields, including math reasoning and code generation.

According to the definition, an AI agent is not necessarily a language model. The system calls the Transformer-based data mixing agent.
The system directly calls the data mixing agent at each data re-weighting step to predict the next mixing weight.

%% file: Appendix/sections/start_state_estimate.tex
\begin{figure*}[htpb]
\centering
\includegraphics[width=12cm,height=9cm]{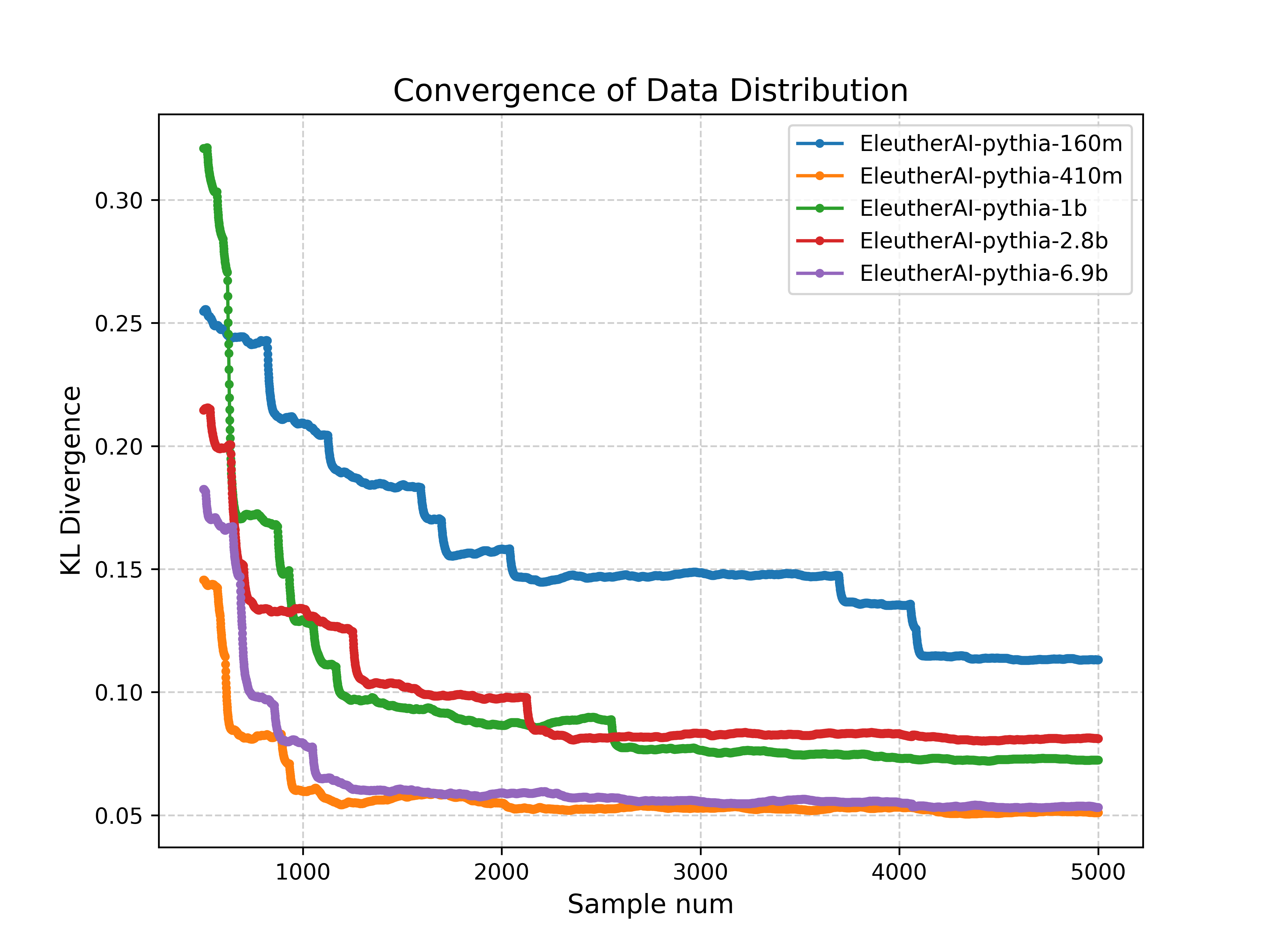}
\caption{The KL divergence between the estimated start state by sampled data from the target model and the ground-truth distribution obtained from the Pile dataset. The results are averages of 5 random runs.}
\label{fig:kl_output}
\end{figure*}

\section{Start State Estimation with Synthetic Data}\label{appn:start_state_estimate}
In scenarios where the data from the source field are unavailable~\citep{grattafiori2024llama,liu2024deepseek}, we explore randomly sampled data from the target model as estimates for the start state. To prove the viability of this method, we experiment on five Pythia models~\citep{biderman2023pythia}, as they are trained on the same open source Pile dataset~\citep{gao2020pile}, where the ground-truth start state can be calculated using the same method as in the paper. 
Specifically, we first randomly sample tokens from the target model simply by pre-pending the start-of-sentence token to start generation with a default temperature of $1.0$. The generated data are then passed through the same Nvidia domain classifier to obtain a domain label. The estimated start state is calculated by normalizing sample numbers on each domain in the generated data. 

To prove the viability of such estimation, we calculate the KL divergence between the estimated start state and the ground-truth distribution obtained from the Pile dataset, and the results are presented in Fig. \ref{fig:kl_output}. As shown, sampling over 2,000 random samples already leads to a KL divergence of less than 0.1 on 4 out of 5 Pythia models, showing a fast convergence rate of the proposed estimation method. The estimated distribution also converges on most models with over 4,000 samples. In addition, larger models such as Pythia-6.9B also show more accurate estimates and a faster convergence rate, possibly due to the higher quality of their generated data, leading to more accurate domain classification. The above results prove the effectiveness of using random samples from the target model as estimates for their start states. This method is further utilized in experiments on the Pythia-1.4B model.

%% file: Appendix/sections/implementation.tex
\section{Implementation Details}
\label{appn:implement_detail}
In this section, we provide more technical implementation details of Data Mixing Agent, including the trajectory sampling algorithm, the evaluation environment feedback design, the data mixing agent training process, and our evaluation process on various settings.

\input{Tables/traj_sampling_algorithm}

\input{Tables/data-mixing-agent-algorithm}

\subsection{Trajectory Sampling}\label{appn:trajectory_sampling}
\paragraph{Algorithm}
The detailed algorithm for the
trajectory sampling process is provided in Algorithm \ref{algo:traj_sampling}. The function $CalculateInductiveScores$ describes the scoring algorithm. This function is designed based on three inductive biases that denote a potentially good distribution:
\begin{itemize}
    \item The data re-weighting distribution at the current step should not deviate significantly from that of the previous step;
    \item As the data re-weighting progresses, the distribution at each step should gradually align more closely with the target distribution;
    \item The distribution at the current step should differ from those at the same step in previously sampled trajectories to encourage diversity.
\end{itemize}
The target distribution is defined as the complement of the start state: probabilities for source-field domains are set to zero, while those for target-field domains are estimated based on their empirical distribution in the target field data. \textbf{This target distribution serves as a soft constraint that encourages trajectories to gradually reduce reliance on source-field data and increase the coverage of target-field data}, thereby accelerating the continual pre-training process and saving computation costs.

\paragraph{Implementation}
During implementation, we use 100B random tokens from the DCLM~\citep{li2024datacomp} as the source field data $S$, and the math split (about 10B tokens) of the Dolmino-mix-1124~\citep{olmo20242} dataset as the target field data $T$. The max data reweighting steps $M=80$, and the reweighting sample number per step $R$ is set to 8K. To ensure inclusion of both high-quality and low-quality trajectories, we run Algorithm \ref{algo:traj_sampling} four times, each with the path sampling number $P=96$ and the threshold $K$ set to 1, 100, 1000, and 10,000, leading to the trajectory set $\mathcal{T}$ with subsets $\mathcal{T}_{top1}$, $\mathcal{T}_{top100}$, $\mathcal{T}_{top1000}$, and $\mathcal{T}_{top10000}$, with 384 trajectories in total.

\subsection{Evaluation Environment Design and Feedback Collection}
During implementation, the environment assesses the general capability of the checkpoints via 500 high-quality general-domain questions and answers from the validation set of the MMLU dataset~\footnote{\url{https://huggingface.co/datasets/cais/mmlu}}. The math reasoning capability is evaluated with 500 random samples from the training split of the MATH dataset~\footnote{\url{https://huggingface.co/datasets/EleutherAI/hendrycks_math}}. 

Leveraging this evaluation environment, we collect feedback data by training a small proxy model $\mathcal{M}_p$ with the LLaMA model structure and 50M parameters on each sampled trajectory from scratch. The model checkpoint is evaluated on this environment at each data reweighting step, resulting in 27,266 feedbacks. Formally, for the $i$-th data mixing distribution $\rho_i\in\tau$, we obtain a tuple $(\rho_i, \mathcal{R}(\mathcal{M}_p^i))$, where $\mathcal{R}(\mathcal{M}_p^i)$ denotes the environment feedback for the model checkpoint $\mathcal{M}_p^i$ after training on the $i$-th domain reweighting step. Notably, the feedback at the start state is obtained with the initialized base proxy model.

\subsection{SFT-based Warming Up}\label{appn:sft}
We first perform Supervised Fine-Tuning (SFT) to reduce the parameter searching space in the reinforcement learning phase. We train the agent from scratch on the high-quality $\mathcal{T}_{top1}$ trajectory subset (obtained during the trajectory sampling process) with a simple MSE loss. At the data reweighting step $t$, the agent is optimized as follows:
\begin{equation}
    \mathcal{L}_{SFT} =  \sum\left(\hat{\rho}_t - f(\tilde{\rho}_0,\tilde{\rho}_1,...,\tilde{\rho}_{t-1}) \right)^2
\end{equation}
where $\hat{\rho}_t$ denotes the ground-truth distribution in step $t$. Notably, before the SFT process, we standardize the environment feedback on each target field across data reweighting steps within all trajectories by forcing their mean value to 0 and standard deviation to 1. This is to regularize the reward space for the agent and avoid out-of-distribution rewards from unseen target models. The feedback for later reinforcement learning and agent inference processes also utilizes this standardization procedure.

\subsection{Conservative Q-Learning}\label{appn:cql}
We select Conservative Q-Learning (CQL)~\citep{kumar2020conservative} as the optimization algorithm. CQL prevents overestimation of Q-values for out-of-distribution actions by encouraging the learned Q-function to be conservative. Specifically, CQL introduces a conservative penalty for the Q-function optimization process with the following training objective:

\begin{small}
\begin{equation}\label{eqn:cql}
\begin{aligned} 
&\underbrace{\mathbb{E}_{(s,a,r,s') \sim \mathcal{D}} \left[ \left( Q(s,a) - \left( r + \gamma \max_{a'} Q(s', a') \right) \right)^2 \right]}_{\text{Bellman error}} \\
&+ \alpha \cdot \underbrace{\left( \mathbb{E}_{s \sim \mathcal{D}} \left[ \mathbb{E}_{a' \sim \mathcal{U(\mathcal{A})}} (Q(s,a')) \right] - \mathbb{E}_{(s,a) \sim \mathcal{D}} [Q(s,a)] \right)}_{\text{Conservative penalty}}
\end{aligned}
\end{equation}
\end{small}

where the first term denotes the standard Bellman error for Q-learning, and the second term denotes a conservative penalty to minimize the Q value expectations for randomly sampled actions from the current state under another distribution $\mathcal{U}(\mathcal{A})$.
The target model and the Q function are then trained in an actor-critic~\citep{sutton1999policy} structure, where the data agent acts as the actor model, optimized with policy gradient to enhance actions that maximize the Q value feedback from the critic model. Another neural network is initialized from scratch as the parameterized Q function, which acts as the critic model to evaluate model actions and optimized via Eqn. \ref{eqn:cql}.

During implementation, we randomly sample fragments $\tau$ (don't have to be full trajectories) from the data mixing trajectory set $\mathcal{T}$. At domain re-weighting step $t$, $s=[\rho_0,\rho_1,...,\rho_{t-1}]$, $a=\rho_t$, and $s'=[\rho_0,\rho_1,...,\rho_t]$. The scalar reward value $r$ is obtained as the gain of a linear combination of environment feedback $\mathcal{R}(\mathcal{M}_p^t)$ compared to that of the last step:
\begin{equation}
r=\sum_{i=1}^{|D|}\lambda_i\mathcal{S}(\mathcal{M}_p^t, D_i)-\sum_{i=1}^{|D|}\lambda_i\mathcal{S}(\mathcal{M}_p^{t-1}, D_i)
\end{equation}
During implementation, we set all coefficients to be equal to encourage balanced consideration of target field capabilities: $\lambda_i=\frac{1}{|D|}$. The critic model $f'$ is parameterized by another single-layer Transformer decoder, followed by a linear layer and sigmoid function to project the representations into a Q-value scalar, with the following inference function:
\begin{equation}
    Q(s, a)= f'(\rho_0,\rho_1,...,\rho_t)
\end{equation}
The agent model and the Q-function model are iteratively optimized in this actor-critic manner until convergence.

The advantage of CQL over SFT is expected. The theoretical foundation of the data mixing agent lies in learning from both actions that enhance and degrade model performance. CQL, as SOTA off-policy RL algorithm, is inherently designed to optimize policies using both positive and negative signals from sampled trajectories. Specifically, it promotes actions associated with high Q-values while suppressing those associated with low Q-values.

In contrast, SFT is an instance of imitation learning, which optimizes the policy by fitting it to high-quality (i.e., “good”) trajectories. By design, SFT cannot exploit informative signals from suboptimal or “bad” trajectories, where the policy should instead learn to avoid certain actions. When incorporated, such trajectories may even introduce noise rather than useful learning signals.
For this reason, we employ SFT only as a warm-up strategy and subsequently apply CQL to fully optimize the policy over all trajectories.

We believe the strong generalization capability of the agent stems from the same intuition that motivated the development of the data mixing agent in the first place: there exists a rich heuristic space for domain reweighting. These heuristics are largely model- and data-agnostic, and can therefore be unified within a compact agent model to guide data mixing trajectories in a principled manner. Our approach differs from prior work in that, rather than relying on manually designed algorithms, we train a model to parameterize these heuristics and enable automated domain reweighting. As a result, we expect our method to achieve comparable or superior generalization performance, consistent with the empirical success of previous heuristic-based approaches.

\subsection{System Pipeline of Data Mixing Agent}\label{appn:pipeline}
Following Google’s definition~\footnote{\url{https://cloud.google.com/discover/what-are-ai-agents}}, an AI agent is a software system that autonomously pursues goals and completes tasks by exhibiting reasoning, planning, and adaptation capabilities. Under this definition, the proposed data mixing agent qualifies as an AI agent. Concretely, it is implemented as a Transformer-based learning system that operates as an independent decision-making module. Its objective is to predict an optimal data mixing recipe for the next domain reweighting step, given the historical training trajectory as input. By encoding past data mixing decisions and their associated outcomes, the agent effectively maintains memory over previous trajectories and reasons over them to plan future data allocation strategies. Moreover, empirical results demonstrate that the agent can learn and adapt across domains, generalizing effectively to diverse target fields such as math reasoning and code generation.

In the overall system pipeline, the data mixing agent is invoked directly at each reweighting step to predict the next set of domain mixing weights, rather than calling a language model for decision making. This design aligns with the general notion of AI agents as decision-making systems that are not necessarily language models themselves.

%% file: Tables/traj_sampling_algorithm.tex
\begin{algorithm*}[htbp]
\caption{Data Mixing Trajectory Sampling with Top-K Inductive Biases}\label{algo:traj_sampling}
\KwIn{Source field Data $S$, Target field data $T$, Path sampling number $P$, Max data reweighting steps $M$, Reweighting sample number per step $R$, Inductive threshold $K$}
\KwOut{Sampled trajectories $\mathcal{T}$}

$D \leftarrow$ GetDomainConfig() \tcp*{Load the domain space based on definitions.}
$\rho_s \leftarrow$ GetStartState($S$, $D$) \tcp*{Estimate start state from source data $S$.}
$\rho_t \leftarrow$ GetTargetState($T$, $D$) \tcp*{Estimate target state from source data $S$.}
$T_M \leftarrow R \cdot M$ \tcp*{Max data samples for each trajectory.}
$\mathcal{T} \leftarrow [\,]$ \tcp*{Initialize empty trajectory list.}

\For{$p \leftarrow 1$ \KwTo $P$}{
    $d \leftarrow 0$, $c \leftarrow 0$, $\rho \leftarrow \rho_s$, $\tau \leftarrow [\rho_s]$ \tcp*{Initialize the current trajectory.}

    \While{$d < M$}{
        $\mathcal{C} \leftarrow [\,]$ \tcp*{Reset candidate list.}

        \For{$i \leftarrow 1$ \KwTo $20000$\tcp*{Repeat the sampling 20,000 times.}} {
            $\rho' \leftarrow$ RandomProbability($|D|$) \tcp*{Randomly sample a distribution.}
            $s \leftarrow$ CalculateInductiveScores($d$, $\rho'$, $\mathcal{T}$, $\tau$, $\rho$, $\rho_t$) \;
            Append $(\rho', s)$ to $\mathcal{C}$ \;
        }

        $\hat{\rho} \leftarrow$ RandomTopK($\mathcal{C}$, $K$) \tcp*{Randomly select from top-$K$ candidates with lowest inductive scores.}
        Append $\hat{\rho}$ to $\tau$ \tcp*{Update current trajectory.}
        $\rho \leftarrow \hat{\rho}$, $d \leftarrow d + 1$ \;
        $c \leftarrow c +$ TargetSamplesCovered($\hat{\rho}$, $R$) \tcp*{Track covered target sample number.}

        \If{$c \ge |T|$}{
            \textbf{break} \tcp*{Early stopping if target data is fully covered.}
        }
    }

    Append $\tau$ to $\mathcal{T}$ \tcp*{Store the current trajectory.}
}
\SetKwFunction{FHeuristic}{CalculateInductiveScores}
\SetKwProg{Fn}{Function}{:}{}
\Fn{\FHeuristic{$d$, $\rho'$, $\mathcal{T}$, $\tau$, $\rho$, $\rho_t$}}{
    $s_c \leftarrow \mathrm{KL}(\rho \,||\, \rho')$ \tcp*{KL divergence between the current and last action.}
    $s_t \leftarrow \mathrm{KL}(\rho_t \,||\, \rho')$ \tcp*{KL divergence between the current action and target state.}
    $s_d \leftarrow 0$ \;

    \If{$|\mathcal{T}| > 0$}{
        $S \leftarrow [\,]$ \tcp*{A set to store the similarities.}
        \ForEach{$\tau' \in \mathcal{T}$}{
            \If{$d < |\tau'|$}{
                $S \leftarrow S \cup \left\{\mathrm{KL}(\tau'[d] \,||\, \rho')\right\}$ \tcp*{Calculate similarities to states in previous trajectories.}
            }
        }
        \If{$|S| > 0$}{
            $s_d \leftarrow \frac{1}{|S|} \sum_{x \in S} x$ \tcp*{Average similarity to previous states}
        }
    }
    \Return $\alpha \cdot s_c + \beta \cdot \sigma\left(\frac{d}{5}\right) \cdot s_t - \gamma \cdot s_d$ \tcp*{Final inductive score}
}
\end{algorithm*}

%% file: Tables/data-mixing-agent-algorithm.tex
\begin{algorithm*}[htbp]
\caption{Continal Pre-training with Data Mixing Agent}\label{algo:data_mixing_agent}
\KwIn{$N$ domains of source data $\{S_1,S_2,...,S_N\}$ and target data $\{T_1,T_2,...,T_N\}$, the agent $f$, Max data reweighting steps $M_{tgt}$, Reweighting sample number per step $R_{tgt}$, the target model $\mathcal{M}_{tgt}$, the evluation environment $\mathcal{E}$.}
\KwOut{The continually pretrained target model checkpoint $\hat{\mathcal{M}}_{tgt}$}

$D \leftarrow$ GetDomainConfig() \tcp*{Load the domain space based on definitions.}
$\rho_s \leftarrow$ GetStartState($S$, $D$) \tcp*{Estimate start state from source data $S$.}
% $\rho_s \leftarrow$ GetStartFeedback($\mathcal{M}_{tgt}$) \tcp*{Estimate start feedback from the target model.}
$\hat{\mathcal{M}}_{tgt} \leftarrow \mathcal{M}_{tgt}$ \tcp*{Initialize the current model checkpoint.}
$\mathcal{T}_{tgt} \leftarrow [\rho_s]$ \tcp*{Initialize trajectory list.}
$reward_{tgt} \leftarrow \varnothing$ \tcp*{Initialize feedback list.}

c$\leftarrow0$

\For{$t \leftarrow 1$ \KwTo $M_{tgt}$}{
    $reward(\hat{\mathcal{M}}_{tgt}) \leftarrow$ GetEnvFeedback($\hat{\mathcal{M}}_{tgt}$, $\mathcal{E}$) \tcp*{Get environment feedback.}
    
    Append $reward(\hat{\mathcal{M}}_{tgt})$ to $reward_{tgt}$ \tcp*{Update current feedback list.}
    
    $reward_{tgt}^s\leftarrow$ std($reward_{tgt}$)\tcp*{Standardize the current feedback list.}
    
    $\{\tilde{\rho}\}\leftarrow$ concat($\mathcal{T}_{tgt}$, $reward_{tgt}^s$)\tcp*{Concat trajectories with the corresponding feedback.}

    $\rho_t\leftarrow f(\tilde{\rho}_1,\tilde{\rho}_2,...,\tilde{\rho}_{t-1})$\tcp*{Obtain domain reweighting distribution from the agent.}

    $\mathcal{B}_t\leftarrow$ sample($\{S_i\}_1^N$,$\{T_i\}_1^N$,$\rho_t$)\tcp*{Sample domain data with the current distribution.}

    Update weights for $\hat{\mathcal{M}}_{tgt}$ with the training loss $\mathcal{L}(\hat{\mathcal{M}}_{tgt}, \mathcal{B}_t)$\;

    Append $\rho_t$ to $\mathcal{T}_{tgt}$\;

    $c \leftarrow c +$ TargetSamplesCovered($\hat{\rho}$, $R$) \tcp*{Track covered target sample number.}

    \If{$c \ge |T|$}{
        \textbf{break} \tcp*{Early stopping if target data is fully covered.}
    }

}
\end{algorithm*}

%% file: Appendix/sections/experiment_set.tex
\section{Experimental Settings}\label{appn:experiment_setting}
\subsection{Target Models}
We aim to rigorously evaluate domain reweighting methods on target models that do not possess math or coding capabilities. Since most open-source models have been optimized on large-scale data from the math reasoning or code generation field, we pre-train three models from scratch, with the same LLaMA3 model architecture~\citep{grattafiori2024llama} of 32 Transformer layers and 3B model parameters, on 100B randomly sampled tokens from the DCLM~\citep{li2024datacomp}, Fineweb-Edu~\citep{penedo2024fineweb}, and Nemotron-CC~\citep{su2024nemotron} dataset, resulting in three target models: \textbf{LLaMA-DCLM}, \textbf{LLaMA-FWE}, \textbf{LLaMA-Nemo}. We also include the \textbf{Pythia-1.4B} model~\citep{biderman2023pythia} to evaluate performance on existing open-source models and scenarios when data from the source field is not directly available.

\subsection{Baseline Methods}
We compare Data Mixing Agent (\textbf{DataAgent$_{RL}$}) with the following baseline methods:
\begin{itemize}
    \item \textbf{Base Model}: direct evaluation of the target models on the benchmarks, reflecting model capabilities before the continual pretraining phase;

    \item \textbf{Naive Training}: continually training the base model on data from the target field without curating any data mixtures from source-field data;

    \item \textbf{RegMix}~\citep{liu2024regmix}: state-of-the-art static domain re-weighting method. It trains large quantities of 1B-sized small proxy models with LLaMA structure (512 models in our implementation) on random domain distributions, then evaluates these models on the target benchmarks. The best data mixing recipe is determined by fitting a regression model to the feedback and selecting distributions that lead to the highest scores. We mostly follow the settings of \citet{wettig2025organize} in implementing the RegMix algorithm;

    \item \textbf{Dynamic Batch Loading (DBL)}~\citep{xia2023sheared}: State-of-the-art dynamic domain reweighting method that modifies the data mixture on the fly. It first estimates a reference optimal loss on each domain reweighting step via the scaling law~\citep{hoffmann2022training}, then computes the excess loss between the current evaluation loss and the optimal loss on each domain, increasing data ratios for slowly learning domains for the next training data batch. We skip the procedure for fitting the scaling function and directly utilize the loss of the base model on the general environment set as the reference loss for the general domain, the loss of the naively trained model on the math environment set as the reference loss for math reasoning, and the loss of the naively trained model on the code environment set as the reference loss for code generation. The DBL uses the same reweighting samples per step as Data Mixing Agent until the target field data is fully covered. We only use DBL as a baseline on the 2-dimensional domain space, due to the lack of evaluation sets for all domains in the 52-dimensional domain space;

    \item \textbf{DataAgent$_{SFT}$}: the data mixing agent model without the reinforcement learning process. The model mostly provides heuristically appropriate trajectories because it's only fine-tuned on the $\mathcal{T}_{top1}$ dataset. We include this baseline method to assess the effectiveness of off-policy optimization with CQL.
\end{itemize}

\subsection{Target Model and Data}\label{appn:target_model_training_data}
For data from the source field, the self-pretrained LLaMA-3B models utilize their corresponding pre-training data, each with 100B tokens. We use randomly sampled data from the Pythia-1.4B model as the source field data for itself, applying the agent in scenarios where the data from the source field is not directly available. Following the method in Appendix \ref{appn:start_state_estimate}, we sample 10B tokens by pre-pending the start-of-sentence token to start generation with a default
temperature of 1.0. The generated data are then filtered through the Nvidia text quality classifier~\footnote{\url{https://huggingface.co/nvidia/quality-classifier-deberta}}, where all data within the "Low quality" class are discarded, resulting in a source field dataset with around 7.7B tokens. For data from the math reasoning field, we select the math split (10B tokens) of the Dolmino-mix-1124 dataset~\footnote{\url{https://huggingface.co/datasets/allenai/dolmino-mix-1124}}, which was used for the mid-training process of the OLMo2 model series~\citep{olmo20242}, including data sources such as TuluMath~\citep{ivison2024unpacking}, MathCoder~\citep{wang2023mathcoder}, and Metamath~\citep{yu2023metamath}. For data from the code generation field, we select the GitHub training split~\footnote{\url{https://huggingface.co/datasets/MBZUAI-LLM/SlimPajama-627B-DC}} of the SlimPajama-DC dataset~\citep{shen2023slimpajama} with 30B tokens.

\subsection{Elaborations on Target Model Selection}
Our experiments aim to validate the proposed data mixing agent as a domain reweighting method rather than to achieve state-of-the-art leaderboard performance. Therefore, absolute performance gaps to SOTA LLMs do not undermine the conclusions.
Specifically, we employ 3B-parameter models pretrained on 100B tokens, whereas strong open-source LLMs (e.g., Llama-3.1-405B-Instruct) use hundreds of billions of parameters and are pretrained on trillions of tokens.
We avoid larger open-source LLMs for three reasons: (1) many LLMs have been heavily optimized for math or code, introducing biases that confound controlled evaluation of domain reweighting; (2) their pretraining data compositions are not publicly available, making them unsuitable for continual pretraining studies; and (3) their large scale (typically>70B parameters) renders continual pretraining prohibitively expensive.

\subsection{Evaluation Benchmarks}
We evaluate target models' general capabilities by evaluating with the lm$\_$eval evaluation library~\footnote{\url{https://github.com/EleutherAI/lm-evaluation-harness}} on the MMLU~\citep{hendrycks2020measuring}, HellaSwag~\citep{zellers2019hellaswag} (Hella.), OpenBookQA~\citep{mihaylov2018can} (OBQA), Winogrande~\citep{sakaguchi2021winogrande} (Wino.), ARC-Challenge~\citep{clark2018think} (ARC-C), PiQA~\citep{bisk2020piqa}, SciQ~\citep{welbl2017crowdsourcing}, and LogiQA~\citep{liu2020logiqa} benchmarks. We evaluate the math reasoning capabilities using the math$\_$lm$\_$eval library~\footnote{\url{https://github.com/ZubinGou/math-evaluation-harness}} on the GSM8K~\citep{cobbe2021training}, MATH~\citep{hendrycks2021measuring}, Minerva~\citep{lewkowycz2022solving}, and MathQA~\citep{amini2019mathqa} benchmarks. We evaluate the code generation capabilities using the eval$\_$plus library~\footnote{\url{https://github.com/evalplus/evalplus}} on the HumanEval~\citep{chen2021evaluating} and MBPP~\citep{austin2021program} benchmarks. The MMLU and GSM8K benchmarks are evaluated with a 5-shot setting, the Minerva benchmark is evaluated in a 4-shot setting. Other benchmarks are evaluated with a zero-shot setting.

\subsection{Target Model Training Setting}
Firstly, the agent is trained based on a 26-dimensional domain definition, leading to a 52-dimensional domain reweighting space. To evaluate on different domain spaces, we further employ the data mixing agent on the original 2-dimensional space based on data sources (source and target). Note that this action does not require the agent to retrain, as its action can be directly converted by summing the 26 probabilities for source/target fields into a single dimension, still preserving a probability distribution. During training, we set the number of reweighting samples per step $R_{tgt}$ to 64K and the maximum data reweighting steps $M_{tgt}$ to 80. We use the same evaluation environment as the Data Mixing Agent training when the target field is math reasoning, and change the MATH validation set to 1,000 random samples from the GitHub validation split of the SlimPajama-DC dataset when the target field is code generation. Due to resource limits and the highly imbalanced distribution of code data in the 26-dimensional domain space, we only train on the 2-dimensional data reweighting space for code generation.

\subsection{Implementation}
We continually pre-train the target model in a distributed manner on 8 nodes with a total of 64 Nvidia A100 GPUs with 40GB of memory. The code for training the target model with data mixing agents is built upon the Megatron-LM framework~\citep{shoeybi2019megatron}. The SFT-based warm-up stage is conducted on the OpenRLHF library~\citep{hu2024openrlhf}. The CQL-based off-policy reinforcement learning framework is built on the d3rlpy~\citep{d3rlpy} library with further modifications to support training on Huggingface Transformer models.

%% file: Appendix/sections/experiment_result.tex
\section{Experimental Results}
\input{Tables/main-results-appn}
\input{Tables/code-results-appn}
\subsection{Full Results and Analysis on Math Reasoning}\label{appn:result_math}
The evaluation results of continual pretraining on the math reasoning target field are shown in Table \ref{tab:main-results-appn}. The data agent and all domain reweighting baseline methods take significantly more GPU training hours than naive training. The main reason is that these domain reweighting methods require a mixture of target-field data and source-field data, while naive training only requires target-field data. By training on more tokens, domain reweighting methods significantly mitigate the catastrophic forgetting problem encountered in naive training, usually outperforming naive training by a large margin on average performance. According to the results, we have the following observations:

\textbf{Naive training significantly improves target model performance on the target field but leads to drastic collapse on the capabilities of the source field.} Compared to the base model, the average math reasoning performance increases by an average of 22.77\% on the four target models, indicating the effectiveness of training on high-quality in-distribution data for the target field. However, the performance on general benchmarks drops by an average of 11.96\%, showing a significant degradation in the source-field model capability. These results further highlight the existence of catastrophic forgetting problems in continual pre-training scenarios, motivating exploration in data mixture and domain reweighting algorithms. While data mixing agent bears an average of 0.57\% gap to naive training in the target domain, it outperforms naive training by an average of 11.99\% in the source domain, showing a significant mitigation of the catastrophic forgetting problem.

\textbf{Domain reweighting algorithms such as RegMix can achieve balanced performance across fields.} According to the results, the RegMix method exhibits a trade-off effect across domains. On the 2-dimensional data reweighting space, it outperforms the base model on math reasoning by an average of 18.47\%, while largely preserving general capabilities with a mere 2.28\% degradation on the corresponding benchmarks. RegMix also outperforms the naive training by 5.03\% on the overall average performance. Similar conclusions can be drawn from the results on the 52-dimensional domain space in Table \ref{tab:nvidia-main-results-appn}. These results show that the catastrophic forgetting problem can be considerably alleviated by carefully curating data mixtures of source and target fields.

\textbf{Data Mixing Agent significantly outperforms other methods in balanced performance across fields.} In Table \ref{tab:2D-main-results-appn}, for the in-distribution LLaMA-DCLM target model, DataAgent$_{RL}$ outperforms the RegMix results on 7 out of 8 general benchmarks and all 4 math benchmarks. It achieves the best average performance 54.04\% and 33.02\% on general/math benchmarks, even outperforming the base model in general ability and the naively trained model on math reasoning. These results prove that DataAgent$_{RL}$ can effectively curate the data mixture to improve both general and math reasoning capabilities. With careful domain reweighting, increasing capability on the target field can further enhance performance on the source field. Overall, DataAgent$_{RL}$ achieves 47.03\% on average, surpassing RegMix by 3.02\%, DBL by 3.53\%, and the base model by 8.88\%. DataAgent$_{RL}$ also outperforms DataAgent$_{SFT}$ by a large margin of 2.08\%. This advantage shows that the empirical guidance presented in the trajectory sampling algorithm is trivial compared to heuristics derived from the broader sampling of data mixing trajectories, underscoring the importance of reinforcement learning with CQL as a crucial step towards capable agents.

\textbf{The capabilities of data mixing agents can generalize across target models, source-field data, and domain spaces without retraining.} Though our data mixing agent is trained on the 52-dimensional data reweighting space with trajectories sampled with the DCLM data, it effectively guides domain reweighting for four target models across 2 domain definitions. For example, in Table \ref{tab:2D-main-results-appn}, DataAgent$_{RL}$ outperforms RegMix by an average of 1.66\%,  DBL by an average of 2.37\% on the two unseen target models: LLaMA-FWE and LLaMA-Nemo, based on the 2-dimensional domain space. In Table \ref{tab:nvidia-main-results-appn}, DataAgent$_{RL}$ outperforms RegMix by an average of 1.41\% based on the 52-dimensional domain space. These results indicate that Data Mixing Agent learns data- and model-agnostic heuristics from the sampled trajectories that can guide domain reweighting on multiple source-field data distributions, which is crucial to the efficiency of this algorithm, as the feedback collection for sampled data trajectories requires considerable computations. With these generalization capabilities, the agent is still expected to perform well in applications to new target models and source-field data without re-training.

\textbf{Data mixing agent is effective in guiding continual pre-training on estimated start state and data mixtures with synthetic source-field data.} We prove this by reweighting domains on the Pythia-1.4B target model with the estimated start state obtained as in Appendix \ref{appn:start_state_estimate} and source-field data obtained as described in Appendix \ref{appn:target_model_training_data}. In math reasoning, DataAgent$_{RL}$ also outperforms RegMix by 0.59\% in Table \ref{tab:2D-main-results-appn} and 3.7\% in Table \ref{tab:nvidia-main-results-appn}. However, the preservation on general capabilities significantly drops, with a 4.69\% and 5.65\% gap on 2-dimensional and 52-dimensional domain spaces.
Overall, DataAgent$_{RL}$ still significantly improves average performance compared to the base model, and outperforms RegMix by 1.05\% in Table \ref{tab:2D-main-results-appn} and 0.61\% in Table \ref{tab:nvidia-main-results-appn}.

\subsection{Full Results and Analysis on Code Generation}\label{appn:result_code}
We evaluate the Data Mixing Agent's generalization to unseen target fields by directly utilizing the agent trained on the math reasoning field to guide domain reweighting for the code generation field. The results are shown in Table \ref{tab:code_result_appn}. We have the following observations:

\textbf{The capabilities of Data Mixing Agent can partially generalize across target fields without retraining.} DataAgent$_{RL}$ achieves the best average performance of 46.3\% and 41.63\% on the LLaMA-DCLM and Pythia-1.4B target models, outperforming the  DBL method by an average of 2.78\% and RegMix method by 2.67\%. These results prove that heuristics learned in the math reasoning field can be partially transferred to the code generation field without modifying the weights of the agent. However, we observe a degradation in DataAgent$_{RL}$'s advantage over the baseline methods in code generation. For example, DataAgent$_{RL}$ outperforms naive training by 6.22\%, while in Table \ref{tab:2D-main-results-appn}, the advantage is 8.52\%. This is mainly due to that applying DataAgent$_{RL}$ to code generation leads to a major 3.18\% drop on general benchmarks compared to math reasoning, which indicates the existence of heuristics that are dependent on the target field and the potential misalignment when converting them to a new target field.

\textbf{The Data Mixing Agent still demonstrates strong generalization to synthetic source-field data and unseen target fields.} This is validated through continual pre-training in the code generation domain using synthetic data from the Pythia-1.4B model. DataAgent$_{RL}$ outperforms RegMix by 2.63\% on general benchmarks, 2.8\% on code benchmarks, and 2.67\% on average. These results highlight the agent’s ability to generalize effectively, enabling its application to scenarios where the source-field data is unavailable and the target model is trained on previously unseen target fields.

\begin{figure*}[htpb]
\centering
\includegraphics[width=12cm,height=7.2cm]{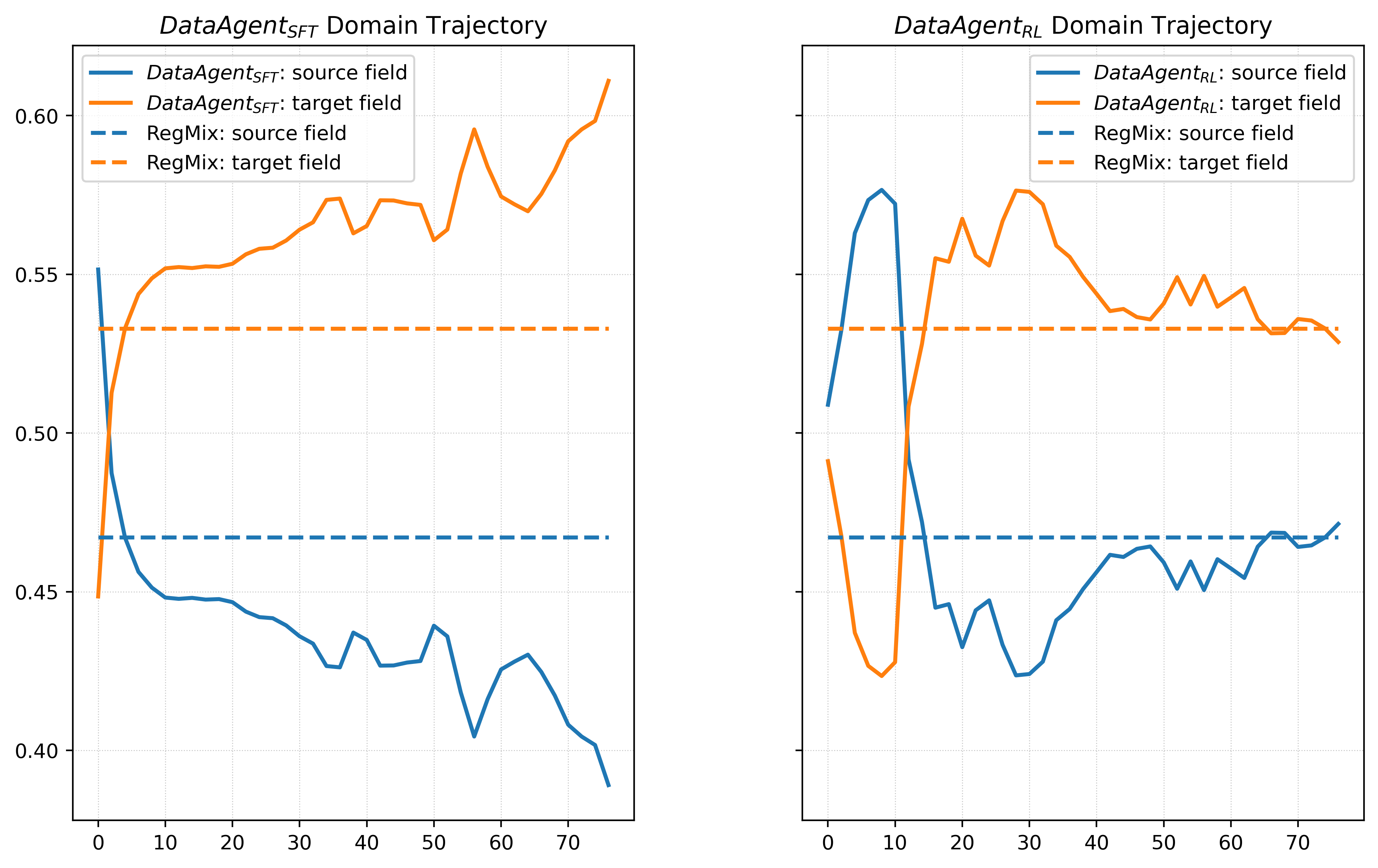}
\caption{The two data mixing agents' output domain reweighting trajectories based on the 2-dimensional domain space, training on the LLaMA-DCLM model and the math reasoning field. The dashed line denotes the optimal domain distributions determined by RegMix.}
\label{fig:data_traj_sft_rl_2D}
\end{figure*}

\begin{figure*}[htpb]
\centering
\includegraphics[width=12cm,height=14.4cm]{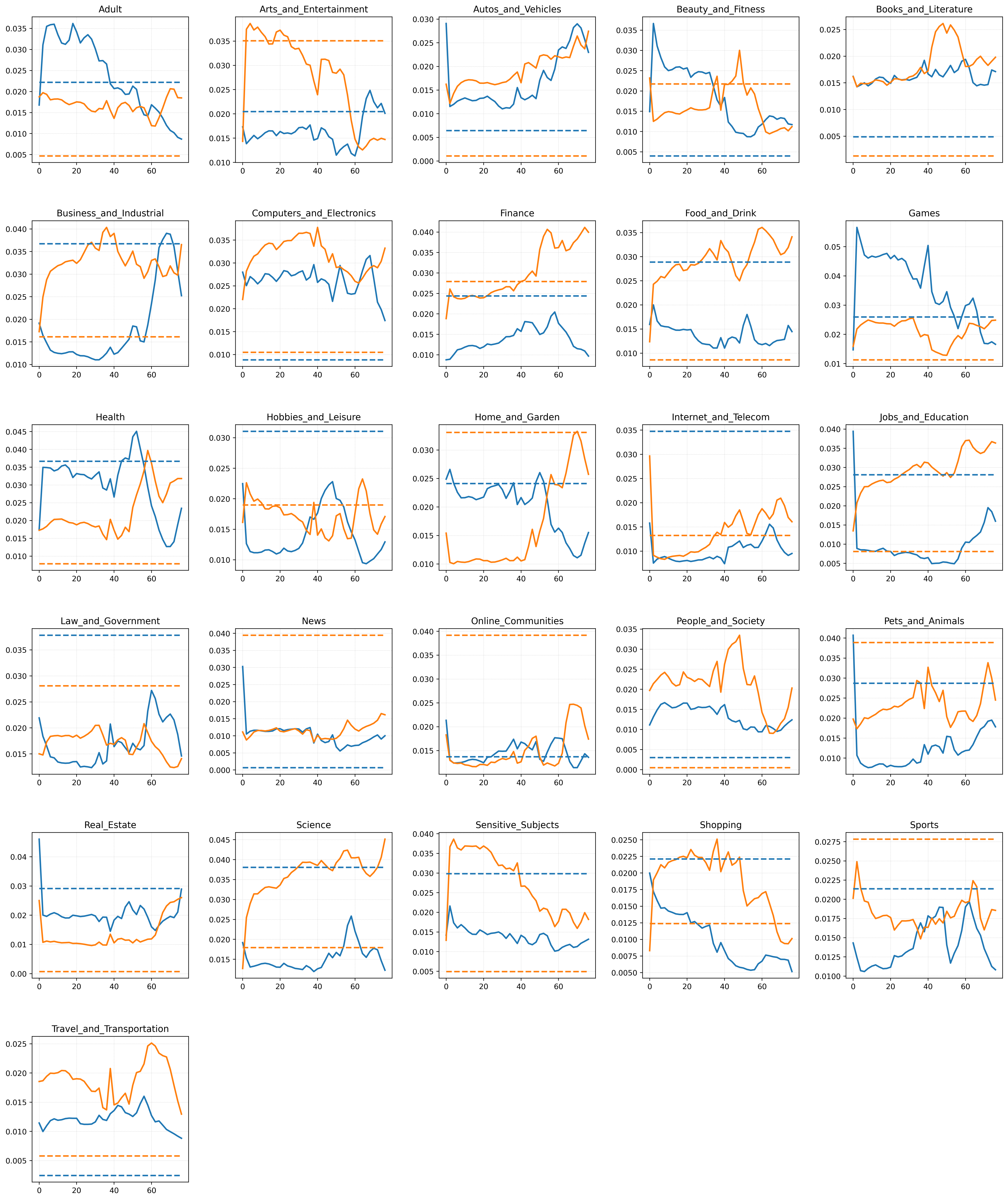}
\caption{DataAgent$_{SFT}$'s domain reweighting trajectories based on the 52-dimensional domain space, training on the LLaMA-DCLM model and the math reasoning field. The legends within each sub-figure are the same as those of Fig. \ref{fig:data_traj_sft_rl_2D}.}
\label{fig:data_traj_sft_nvidia}
\end{figure*}

\begin{figure*}[htpb]
\centering
\includegraphics[width=12cm,height=14.4cm]{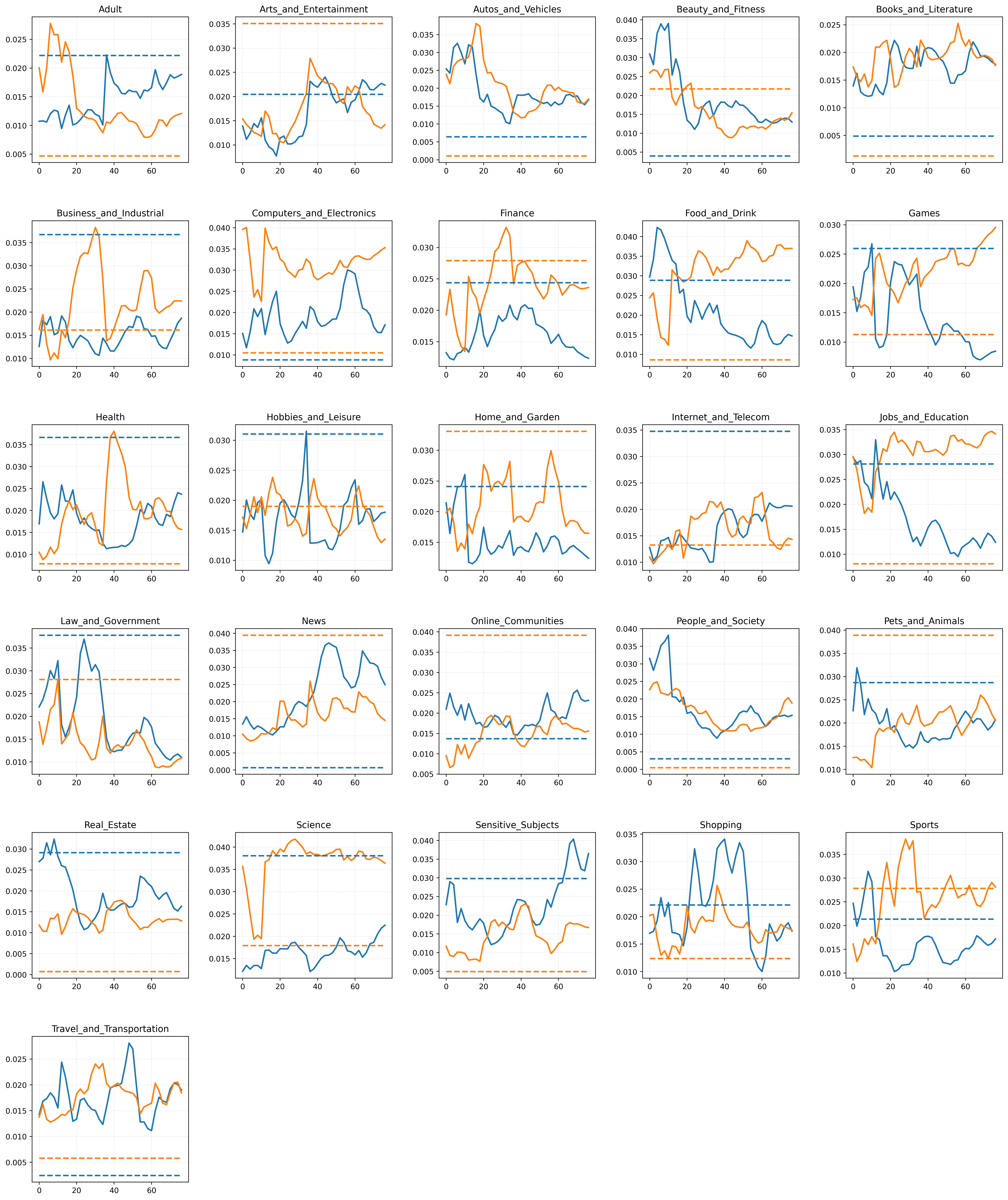}
\caption{DataAgent$_{RL}$'s domain reweighting trajectories based on the 52-dimensional domain space, training on the LLaMA-DCLM model and the math reasoning field. The legends within each sub-figure are the same as those of Fig. \ref{fig:data_traj_sft_rl_2D}.}
\label{fig:data_traj_rl_nvidia}
\end{figure*}

\subsection{Full Analysis on Domain Reweighting Trajectories}\label{appn:result_traj_analysis}
In this section, we showcase the domain reweighting process guided by Data Mixing Agent to train the LLaMA-DCLM model on the math reasoning field, aiming to provide more intuitions on its actions based on the heuristics and feedback. The trajectories on the 2-dimensional and 52-dimensional domain spaces are provided in Fig. \ref{fig:data_traj_sft_rl_2D}, Fig. \ref{fig:data_traj_sft_nvidia}, and Fig. \ref{fig:data_traj_rl_nvidia}. We have the following observations:

\textbf{Data Mixing Agents follow a less-to-more trend when adapting the target field data along the data mixing trajectory, but DataAgent$_{RL}$ adopts a more fine-grained approach to achieve superior performance.} In Fig. \ref{fig:data_traj_sft_rl_2D}, both the DataAgent$_{RL}$ and DataAgent$_{SFT}$ models show an overall trend to increase data from the target field and decrease data from the source field, but with different strategies. DataAgent$_{SFT}$ shows a radical trend towards more target field data, increasing the DCLM data ratio almost monotonically from about 45\% to over 60\% during continual pre-training. DataAgent$_{RL}$ adopts a more conservative three-stage strategy: 
\begin{itemize}
    \item \textbf{Early warm-up stage}: the agent prioritizes source field data to stabilize training;
    \item \textbf{Mid-training stage}: the agent rapidly increases the use of target field data to enhance performance on the target capability;
    \item \textbf{Final stage}: the agent gradually reintroduces more source field data, with the data distribution stabilizing around the optimal weights identified by RegMix.
\end{itemize}
As shown in Table \ref{tab:2D-main-results-appn}, the superior performance of DataAgent$_{RL}$ on both general and math reasoning benchmarks proves the advantage of its subtle domain reweighting strategy. This performance gap between Data Mixing Agents is mainly due to the comprehensive modeling of the heuristic space during reinforcement learning. DataAgent$_{SFT}$ is only fine-tuned on the $\mathcal{T}_{top1}$ trajectories, which mostly model the inductive biases from the $CalculateInductiveScores$ function. DataAgent$_{RL}$ is further optimized on a broad range of trajectories via reinforcement learning, including $\mathcal{T}_{top1}$, $\mathcal{T}_{top100}$, $\mathcal{T}_{top1000}$, and $\mathcal{T}_{top10000}$, with contrastive supervision signals to increase probabilities of actions that improve overall performance and avoid actions that hurt performance measured by the environment feedback. The visualization of the 52-dimensional domain reweighting trajectories further strengthens the above arguments. In Fig. \ref{fig:data_traj_sft_nvidia}, the DataAgent$_{SFT}$ organizes the target field data from about 60\% of the domains to be almost monotonically increasing along the domain reweighting trajectory, while in Fig. \ref{fig:data_traj_rl_nvidia}, the DataAgent$_{RL}$ model introduces more complicated reweighting strategies on about 80\% of the domains.

\textbf{Data Mixing Agents learn heuristics and perform actions that correspond to human intuitions on the target capabilities.} Our work uses the MMLU evaluation set to represent the general capabilities in the environment. \citet{wettig2025organize} summarized the top-3 domains that benefit the MMLU performance: $Science\&Tech.$, $Health$, and $Politics$. In Fig. \ref{fig:data_traj_rl_nvidia}, we observe a significant uplift of the target data distributions in the corresponding domains compared to the RegMix domain distributions: $Science$, $Health$, and $People\&Society$.  \citet{wettig2025organize} also enumerated domains that can hurt performance on MMLU, such as $Fashion\&Beauty$, while DataAgent$_{RL}$ also conveys an explicit down-sampling process in the $Beauty\&Fitness$ domain. These observations further ensure the effectiveness of the learned heuristics, encouraging the discovery of more heuristics via the agent's trajectories. For example, DataAgent$_{RL}$ continuously reduces data from both source and target fields in the $Pets\&Animals$ domain, possibly indicating its lack of importance in enhancing either general or math reasoning capabilities.

\begin{figure*}[htpb]
\centering
\includegraphics[width=12cm,height=5cm]{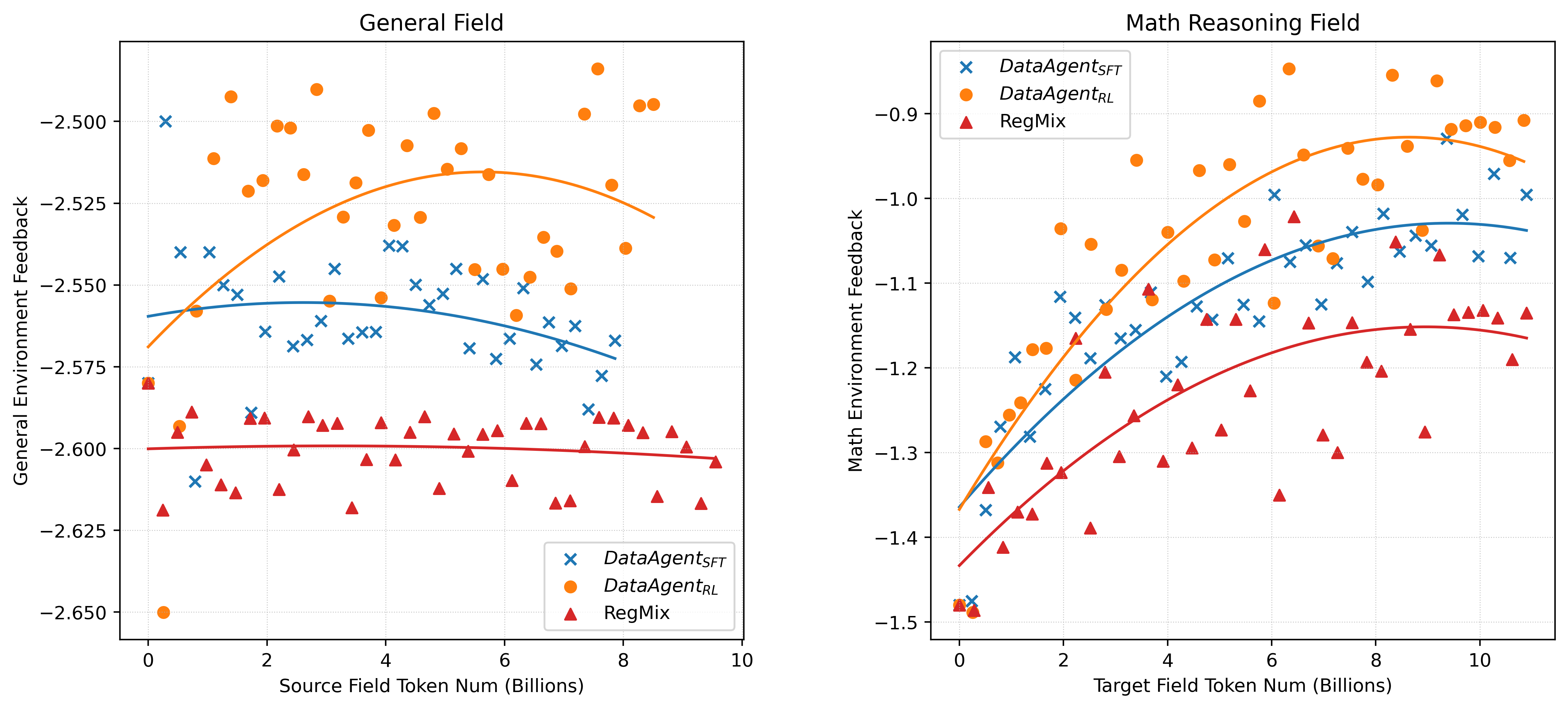}
\caption{The performance dynamics of the target model on the evaluation environment with increasing training data (measured in Billion tokens) on the corresponding field. We set a total training budget of 10.5B tokens, but DataAgent$_{RL}$ triggers an early stopping at 9.96B tokens, and DataAgent$_{SFT}$ triggers an early stopping at 9.43B tokens.}
\label{fig:data_efficiency_2D}
\end{figure*}

\subsection{Full Analysis on Data Efficiency}\label{appn:result_efficiency}
We explore how efficiently the Data Mixing Agents leverage the source and target field data to improve or preserve model capabilities in the corresponding fields. Training on the mixture of DCLM-100B and the math split of Dolmino-mix-1124 datasets, we record the performance dynamics of the LLaMA-DCLM target model on the general/math evaluation environment with increasing training data (measured in Billion tokens) on the general/math reasoning field. The results are shown in Fig. \ref{fig:data_efficiency_2D}. We have the following observations:

\textbf{Data Mixing Agents leverage general field data more efficiently than RegMix, better preserving model capabilities in the source field.} According to the visualization in the general field, the agent methods obtain higher general feedback values from the environment at most token budgets for the source field. The capability measurement for RegMix fluctuates around -2.6, while both data mixing agent models maintain the feedback over -2.575. DataAgent$_{RL}$ further outperforms DataAgent$_{SFT}$ in most cases, with feedback values fluctuating around -2.525, which provides evidence for the heuristics learned during reinforcement learning in preserving the general capabilities. Notably, DataAgent$_{RL}$ shows significantly higher variance in feedback values along the domain reweighting trajectory than both DataAgent$_{SFT}$ and RegMix, reflecting its more active strategies in adjusting domain reweighting distributions to improve source field capabilities. Its final superior performance on MMLU and the average of general benchmarks (as shown in Table \ref{tab:2D-main-results-appn}) further indicates the effectiveness of such strategies.

\textbf{Data Mixing Agents leverage data from the math reasoning field more efficiently than RegMix, resulting in greater improvements in the target field performance.} As presented in the visualization of the math reasoning field, though all methods show logarithmic-scale improvements from the math reasoning environment, the data mixing agent methods show a faster momentum in increasing general feedback values from the environment at most token budgets for the target field. RegMix performance stabilizes around -1.2 while both data mixing agent methods achieve performance over -1.1. These results show that our method can better arrange the continual pre-training data to improve model capability in the target field. DataAgent$_{RL}$ also outperforms DataAgent$_{SFT}$ with the optimized feedback values over -1.0. The leading performance of DataAgent$_{RL}$ on both general and math reasoning fields proves its effectiveness in coordinating the source and target field data to improve performance on multiple target capabilities.

\textbf{Data Mixing Agents achieve balanced continual pre-training performance with less reliance on data from the source field.} As described in Fig. \ref{fig:data_efficiency_2D}, while we set a total training budget of 21B tokens, DataAgent$_{RL}$ triggers an early stopping at 19.92B tokens, and DataAgent$_{SFT}$ triggers an early stopping at 18.86B tokens, due to the exhaustion of the target field data. These results show that the data mixing agent can achieve superior performance than RegMix on both the general and math reasoning fields while relying on 2.14B fewer tokens in the source field, further proving the efficiency of their domain reweighting process.

\subsection{Consistency between Rewards and Model Performance}
To examine whether the proposed environment feedback provides a consistent and reliable signal for downstream model performance, we analyze multiple checkpoints along the LLaMA-DCLM training trajectory in the math domain. Specifically, we evaluate five checkpoints on different training stages (with 3.8B to 19.92B trained tokens) and report their corresponding environment feedback signals (denoted as $Env_{general}$ and $Env_{math}$) and average benchmark performance on general-domain and math tasks (denoted as $bench\_general\_avg$ and $bench\_math\_avg$). 

\begin{table*}[t]
\centering
\begin{tabular}{lccccc}
\toprule
Training Tokens & $Env_{general}$ & $Env_{math}$ & bench\_general\_avg & bench\_math\_avg \\
\midrule
3.8B   & -2.558 & -1.486 & 52.96 & 10.02 \\
7.64B  & -2.564 & -1.058 & 51.08 & 20.70 \\
12.84B & -2.591 & -0.896 & 48.00 & 32.95 \\
16.52B & -2.473 & -0.950 & 58.22 & 23.10 \\
19.92B & -2.500 & -0.902 & 54.04 & 33.02 \\
\bottomrule
\end{tabular}
\caption{Environment feedback signals and benchmark performance at different training stages in the math domain, along the LLaMA-DCLM trajectory.}
\label{tab:env_feedback_correlation}
\end{table*}

The results are shown in Table \ref{tab:env_feedback_correlation}. Across these checkpoints, we observe a strong monotonic relationship between the environment feedback and final benchmark performance. The Pearson's correlation coefficient between $Env_{general}$ and bench\_general\_avg is 0.943, while the correlation between $Env_{math}$ and bench\_math\_avg reaches 0.919. These high correlations indicate that the environment feedback closely tracks the model’s eventual evaluation accuracy, validating its consistency as a training signal. This result demonstrates that the reward used by the data mixing agent is well aligned with downstream task performance, thereby supporting the reliability and effectiveness of our method.

%% file: Tables/main-results-appn.tex
\begin{table*}[htbp]
\centering
\caption{The evaluation results of continual pretraining on the math reasoning target field, reflected on 12 benchmarks. We also separately report the average results on general benchmarks, math reasoning benchmarks, and all benchmarks.}\label{tab:main-results-appn}

% ---------- 2D dimension results ----------
\begin{subtable}{\textwidth}
\centering
\caption{Model performances on the 2-dimensional data reweighting space based on data sources.}\label{tab:2D-main-results-appn}
\resizebox{1.\textwidth}{!}{
\begin{tabular}{lc|ccccccccc|ccccc}
\toprule
\multirow{2}{*}{\textbf{Method}} & \multirow{2}{*}{\textbf{Avg.}} & \multicolumn{9}{c}{\textbf{General Benchmarks}} & \multicolumn{5}{c}{\textbf{Math Benchmarks}} \\
& & MMLU & Hella. & OBQA & Wino. & ARC-C & PiQA & SciQ & LogiQA & Avg. & GSM8K & Minerva & MATH & MathQA & Avg. \\
\midrule
\multicolumn{15}{c}{\textbf{LLaMA-DCLM}}\\
Base Model & 38.15 & 34.5 & \textbf{64.5} & 37.0 & 61.56 & 36.69 & \textbf{75.84} & 84.2 & 28.11 & 52.8 & 2.55 & 4.1 & 4.22 & 24.52 & 8.85 \\
Naive Training & 38.51 & 27.11 & 37.0 & 28.2 & 54.22 & 28.58 & 60.28 & 68.7 & 26.88 & 41.37 & 59.21 & \textbf{16.16} & \textbf{22.85} & 32.96 & 32.80 \\
RegMix & 44.01 & 30.42 & 59.72 & 36.6 & 61.72 & 34.73 & 73.88 & 85.1 & 28.73 & 51.36 & 55.87 & 11.5 & 17.7 & 32.16 & 29.31\\
DBL & 43.50 & 29.0 & 55.66 & 35.0 & \textbf{63.64} & 32.11 & 72.5 & \textbf{88.42} & 28.89 & 50.65 & 56.22 & 12.2 & 18.24 & 30.14 & 29.2\\
DataAgent$_{SFT}$ & 44.95 & 33.81 & 60.23 & 34.2 & 60.43 & 36.26 & 73.28 & 87.3 & 29.33 & 51.86 & 57.84 & 12.7 & 21.3 & 32.73 & 31.14\\
DataAgent$_{RL}$ & \textbf{47.03} & \textbf{34.06} & 63.38 & \textbf{42.14} & 62.35 & \textbf{36.92} & 74.85 & 87.89 & \textbf{30.29} & \textbf{54.04} & \textbf{59.24} & 14.8 & 22.75 & \textbf{35.3} & \textbf{33.02}\\
\midrule
\multicolumn{15}{c}{\textbf{LLaMA-FWE}}\\
Base Model & 37.65 & 34.47 & 60.52 & 37.8 & 57.77 & 40.53 & 74.21 & 85.4 & \textbf{28.26} & 52.37 & 2.5 & 2.52 & 4.06 & 23.72 & 8.2 \\
Naive Training & 38.51 & 27.25 & 37.03 & 28.4 & 53.51 & 26.96 & 61.1 & 69.6 & 28.11 & 41.5 & \textbf{58.91} & 12.58 & \textbf{24.7} & \textbf{33.94} & \textbf{32.53}\\
RegMix & 43.83 & 32.83 & 60.2 & 36.45 & 54.51 & 37.17 & 71.72 & 86.5 & 28.0 & 50.92 & 55.9 & 12.2 & 20.35 & 30.1 & 29.64 \\
DBL & 43.92 & 31.27 & 56.58 & \textbf{40.7} & 56.27 & 38.32 & 71.18 & \textbf{88.26} & 26.0 & 51.07 & 54.2 & 12.05 & 20.72 & 31.52 & 29.62\\
DataAgent$_{SFT}$ & 45.23 & \textbf{34.65} & \textbf{60.83} & 38.28 & 59.3 & \textbf{40.8} & \textbf{74.6} & 85.6 & 26.96 & \textbf{52.63} & 56.26 & 12.19 & 21.92 & 31.32 & 30.42\\
DataAgent$_{RL}$ & \textbf{45.48} & 33.78 & 60.44 & 38.8 & \textbf{59.59} & 38.89 & 73.12 & 84.9 & 27.49 & 52.13 & 58.07 & \textbf{13.46} & 23.96 & 33.28 & 32.19\\
\midrule
\multicolumn{15}{c}{\textbf{LLaMA-Nemo}}\\
Base Model & 38.22 & 34.22 & 64.51 & 37.6 & 59.12 & 36.26 & \textbf{75.57} & \textbf{88.4} & 26.73 & 52.8 & 2.5 & 4.85 & 5.3 & 23.62 & 9.07\\
Naive Training & 37.86 & 27.06 & 37.4 & 28.0 & 52.88 & 27.05 & 59.74 & 68.6 & 25.19 & 40.74 & \textbf{59.05} & \textbf{12.3} & \textbf{24.52} & \textbf{32.53} & \textbf{32.1}\\
RegMix & 44.13 & \textbf{35.03} & \textbf{65.15} & 35.9 & 59.83 & 36.45 & 72.85 & 86.2 & 28.88 & 52.54 & 49.39 & 10.7 & 19.23 & 30.0 & 27.33\\
DBL & 42.63 & 33.2 & 64.82 & 34.0 & \textbf{62.06} & \textbf{39.23} & 70.79 & 78.11 & 24.0 & 50.77 & 45.91 & 10.74 & 20.01 & 28.74 & 26.35\\
DataAgent$_{SFT}$ & 44.12 & 34.06 & 64.25 & 39.04 & 60.4 & 38.1 & 74.17 & 86.75 & 29.16 & 53.24 & 47.81 & 9.07 & 17.8 & 28.86 & 25.89\\
DataAgent$_{RL}$ & \textbf{45.8} & 34.27 & 63.95 & \textbf{39.8} & 61.58 & 38.74 & 74.49 & 86.9 & \textbf{29.95} & \textbf{53.71} & 54.28 & 10.83 & 22.94 & 31.85 & 29.98\\
\midrule
\multicolumn{15}{c}{\textbf{Pythia-1.4B}}\\
Base Model & 33.47 & \textbf{30.74} & \textbf{52.0} & 33.2 & \textbf{57.3} & \textbf{28.33} & \textbf{70.89} & \textbf{79.3} & \textbf{27.5} & \textbf{47.41} & 1.67 & 4.39 & 2.1 & 14.16 & 5.58\\
Naive Training & 31.06 & 25.8 & 26.76 & 18.8 & 41.93 & 21.81 & 52.95 & 62.2 & 21.11 & 33.92 & \textbf{48.98} & \textbf{10.16} & \textbf{14.64} & \textbf{27.52} & \textbf{25.33}\\
RegMix & 34.07 & 29.37 & 48.3 & \textbf{33.4} & 43.88 & 21.16 & 65.7 & 64.46 & 25.2 & 41.43 & 40.2 & 6.92 & 7.93 & 22.28 & 19.33\\
DataAgent$_{SFT}$ & 33.6 & 30.66 & 45.23 & 26.6 & 43.83 & 22.25 & 61.94 & 72.2 & 26.88 & 41.2 & 37.1 & 7.19 & 7.85 & 21.45 & 18.4\\
DataAgent$_{RL}$ & \textbf{35.12} & 30.0 & 48.5 & 30.6 & 41.17 & 25.66 & 65.54 & 75.8 & 24.49 & 42.72 & 40.26 & 7.73 & 8.75 & 22.94 & 19.92\\
\bottomrule
\end{tabular}}
\end{subtable}

\vspace{0.8em}

% ---------- Nvidia results ----------
\begin{subtable}{\textwidth}
\centering
\caption{Model performances on the 52-dimensional data reweighting space based on the Nvidia domain classifier.}\label{tab:nvidia-main-results-appn}
\resizebox{1.\textwidth}{!}{
\begin{tabular}{lc|ccccccccc|ccccc}
\toprule
\multirow{2}{*}{\textbf{Method}} & \multirow{2}{*}{\textbf{Avg.}} & \multicolumn{9}{c}{\textbf{General Benchmarks}} & \multicolumn{5}{c}{\textbf{Math Benchmarks}} \\
& & MMLU & Hella. & OBQA & Wino. & ARC-C & PiQA & SciQ & LogiQA & Avg. & GSM8K & Minerva & MATH & MathQA & Avg. \\
\midrule
\multicolumn{15}{c}{\textbf{LLaMA-DCLM}}\\
Base Model & 38.15 & \textbf{34.5} & \textbf{64.5} & 37.0 & 61.56 & 36.69 & \textbf{75.84} & 84.2 & 28.11 & 52.8 & 2.55 & 4.1 & 4.22 & 24.52 & 8.85 \\
Naive Training & 38.51 & 27.11 & 37.0 & 28.2 & 54.22 & 28.58 & 60.28 & 68.7 & 26.88 & 41.37 & \textbf{59.21} & 16.16 & 22.85 & \textbf{32.96} & \textbf{32.80} \\
RegMix & 44.67 & 34.38 & 62.17 & 38.2 & 61.93 & \textbf{36.95} & 74.97 & 87.5 & 29.49 & 53.19 & 55.78 & 10.36 & 15.75 & 28.67 & 27.64 \\
DataAgent$_{SFT}$ & 45.75 & 34.47 & 63.36 & 40.6 & 62.35 & 35.87 & 74.32 & \textbf{89.2} & \textbf{29.96} & 53.77 & 56.77 & 11.12 & 18.76 & 32.3 & 29.74\\
DataAgent$_{RL}$ & \textbf{46.84} & 32.99 & 62.64 & \textbf{41.6} & \textbf{63.98} & 36.64 & 73.5 & 89.1 & 31.5 & \textbf{53.99} & 59.04 & \textbf{16.48} & \textbf{22.9} & 31.72 & 32.54 \\
\midrule
\multicolumn{15}{c}{\textbf{LLaMA-FWE}}\\
Base Model & 37.65 & 34.47 & 60.52 & 37.8 & 57.77 & \textbf{40.53} & 74.21 & 85.4 & 28.26 & \textbf{52.37} & 2.5 & 2.52 & 4.06 & 23.72 & 8.2 \\
Naive Training & 38.51 & 27.25 & 37.03 & 28.4 & 53.51 & 26.96 & 61.1 & 69.6 & 28.11 & 41.5 & 58.91 & 12.58 & \textbf{24.7} & \textbf{33.94} & 32.53\\
RegMix & 43.78 & \textbf{35.64} & 57.64 & \textbf{39.4} & 57.56 & 38.2 & 67.4 & 85.03 & 29.34 & 51.28 & 53.68 & 10.84 & 20.35 & 30.0 & 28.72\\
DataAgent$_{SFT}$ & 45.21 & 34.27 & 61.2 & 37.95 & 58.57 & 40.28 & \textbf{75.17} & 85.8 & 28.23 & 52.68 & 54.5 & 13.27 & 22.17 & 31.1 & 30.26\\
DataAgent$_{RL}$ & \textbf{45.55} & 32.55 & 58.64 & 35.8 & \textbf{59.42} & 39.76 & 73.34 & \textbf{87.2} & \textbf{29.35} & 52.01 & \textbf{58.96} & \textbf{14.68} & 23.62 & 33.32 & \textbf{32.65}\\
\midrule
\multicolumn{15}{c}{\textbf{LLaMA-Nemo}}\\
Base Model & 38.22 & 34.22 & 64.51 & 37.6 & 59.12 & 36.26 & \textbf{75.57} & 88.4 & 26.73 & 52.8 & 2.5 & 4.85 & 5.3 & 23.62 & 9.07\\
Naive Training & 37.86 & 27.06 & 37.4 & 28.0 & 52.88 & 27.05 & 59.74 & 68.6 & 25.19 & 40.74 & \textbf{59.05} & \textbf{12.3} & \textbf{24.52} & \textbf{32.53} & \textbf{32.1}\\
RegMix & 44.86 & \textbf{35.68} & \textbf{66.4} & 41.4 & \textbf{61.56} & 35.82 & 72.4 & \textbf{87.53} & 29.34 & \textbf{53.77} & 48.17 & 10.91 & 19.03 & 30.04 & 27.04\\
DataAgent$_{SFT}$ & 45.16 & 34.8 & 64.01 & 38.2 & 60.73 & 36.9 & 73.92 & 87.3 & \textbf{29.37} & 53.15 & 52.5 & 10.7 & 20.77 & 32.71 & 29.17\\
DataAgent$_{RL}$ & \textbf{45.9} & 33.89 & 62.7 & \textbf{42.6} & 59.82 & \textbf{40.0} & 74.86 & 84.34 & 28.29 & 53.31 & 56.78 & 11.82 & 23.61 & 32.03 & 31.06\\
\midrule
\multicolumn{15}{c}{\textbf{Pythia-1.4B}}\\
Base Model & 33.47 & 30.74 & \textbf{52.0} & 33.2 & \textbf{57.3} & \textbf{28.33} & \textbf{70.89} & \textbf{79.3} & 27.5 & \textbf{47.41} & 1.67 & 4.39 & 2.1 & 14.16 & 5.58\\
Naive Training & 31.06 & 25.8 & 26.76 & 18.8 & 41.93 & 21.81 & 52.95 & 62.2 & 21.11 & 33.92 & \textbf{48.98} & \textbf{10.16} & \textbf{14.64} & \textbf{27.52} & \textbf{25.33}\\
RegMix & 34.64 & 29.6 & 47.02 & \textbf{34.78} & 46.32 & 24.76 & 62.83 & 69.3 & 27.03 & 42.71 & 38.28 & 7.24 & 7.51 & 21.1 & 18.53\\
DataAgent$_{SFT}$ & 34.01 & \textbf{30.95} & 46.93 & 32.2 & 43.93 & 21.23 & 62.2 & 68.8 & \textbf{27.88} & 41.77 & 38.84 & 7.19 & 7.3 & 20.68 & 18.5\\
DataAgent$_{RL}$ & \textbf{35.25} & 28.8 & 46.97 & 28.8 & 43.61 & 25.91 & 65.3 & 71.31 & 23.36 & 41.76 & 45.44 & 8.1 & 10.01 & 25.38 & 22.23\\
\bottomrule
\end{tabular}}
\end{subtable}
\end{table*}

%% file: Tables/code-results-appn.tex
\begin{table*}[htbp]
\centering
\caption{The evaluation results of continual pretraining on the code generation target field, reflected on 10 benchmarks. The data is reweighted based on the 2-dimensional domain space.}
\label{tab:code_result_appn}
\resizebox{1.\textwidth}{!}{
\begin{tabular}{lc|ccccccccc|ccc}
\toprule
\multirow{2}{*}{\textbf{Method}} & \multirow{2}{*}{\textbf{Avg.}} & \multicolumn{9}{c}{\textbf{General Benchmarks}} & \multicolumn{3}{c}{\textbf{Code Benchmarks}} \\
& & MMLU & Hella. & OBQA & Wino. & ARC-C & PiQA & SciQ & LogiQA & Avg. & HumanEval & MBPP & Avg. \\
\midrule
\multicolumn{14}{c}{\textbf{LLaMA-DCLM}}\\
Base Model & 44.52 & \textbf{34.5} & \textbf{64.5} & 37.0 & \textbf{61.56} & \textbf{36.69} & 75.84 & \textbf{84.2} & 28.11 & \textbf{52.8} & 8.6 & 14.2 & 11.4 \\
Naive Training & 40.08 & 27.6 & 42.96 & 29.37 & 53.1 & 24.76 & 70.5 & 62.95 & 24.46 & 41.96 & \textbf{27.3} & \textbf{37.8} & \textbf{32.55} \\
RegMix & 44.85 & 31.22 & 57.13 & 33.4 & 57.43 & 29.05 & \textbf{76.1} & 82.09 & 29.33 & 49.47 & 21.1 & 31.6 & 26.35 \\
DBL & 43.52 & 31.7 & 55.7 & 35.2 & 53.8 & 31.28 & 68.58 & 78.26 & \textbf{29.8} & 48.04 & 20.6 & 30.3 & 25.45\\
DataAgent$_{SFT}$ & 45.07 & 32.69 & 63.43 & 35.0 & 49.88 & 33.46 & 72.85 & 78.66 & 29.61 & 49.45 & 22.4 & 32.8 & 27.6 \\
DataAgent$_{RL}$ & \textbf{46.3} & 33.84 & 63.79 & \textbf{37.8} & 57.3 & 35.05 & 73.2 & 78.57 & 27.34 & 50.86 & 22.0 & 34.1 & 28.05 \\
\midrule
\multicolumn{14}{c}{\textbf{Pythia-1.4B}}\\
Base Model & 38.91 & \textbf{30.74} & \textbf{52.0} & \textbf{33.2} & \textbf{57.3} & \textbf{28.33} & \textbf{70.89} & \textbf{79.3} & \textbf{27.5} & \textbf{47.41} & 4.9 & 4.9 & 4.9\\
Naive Training & 35.14 & 26.03 & 38.48 & 24.4 & 46.61 & 21.59 & 59.87 & 56.73 & 20.58 & 36.79 & \textbf{24.7} & \textbf{32.4} & \textbf{28.55}\\
RegMix & 38.96 & 28.2 & 47.66 & 28.6 & 52.09 & 26.76 & 64.3 & 69.49 & 24.78 & 42.74 & 20.7 & 27.0 & 23.85 \\
DataAgent$_{SFT}$ & 40.82 & 30.1 & 47.93 & 30.44 & 54.46 & 26.76 & 69.87 & 73.11 & 24.87 & 44.69 & 30.1 & 29.1 & 25.35 \\
DataAgent$_{RL}$ & \textbf{41.63} & 29.9 & 46.31 & 31.8 & 56.9 & 27.35 & 69.2 & 78.26 & 23.27 & 45.37 & 23.3 & 30.0 & 26.65 \\
\bottomrule
\end{tabular}}
\end{table*}

%% file: Sections/related_work.tex
\section{Related Work}\label{appn:related_work}
\subsection{Continual Pre-training}
Continual pre-training is an effective and efficient method for adapting LLMs to new target fields where the pre-training data do not align well, such as knowledge-intensive and complex-reasoning tasks. In math reasoning, DeepSeekMath~\citep{shao2024deepseekmath} was initialized with the DeepSeekCoder~\citep{guo2024deepseek} models and continually trained on 500B tokens of high-quality math-related data. In code generation, the Qwen2.5-Coder~\citep{hui2024qwen2} is based on the Qwen2.5 foundation model and continuously trained on 3.64T tokens of data in the code field. Continual pre-training is also used in other fields such as finance~\citep{xie2024finben}, system research~\citep{lin2025sigma}, and medicine~\citep{tu2024towards}.

The catastrophic forgetting problem is widely encountered in continual pre-training works~\citep{hui2024qwen2,lin2025sigma,luo2023empirical,yang2024qwen2}. Existing works mostly curate mixtures of data from the target field and data from the original field to obtain balanced performance. For example, Qwen2.5-Coder manually determined an optimal data mixing recipe of 7:2:1 in code data, text data, and math data for the Qwen2.5-Coder training dataset, leading to over 20\% improvement in average performance on multiple fields compared to training solely on code data.

\subsection{Data Re-weighting in Pre-training}
Domain reweighting is an emerging research field that aims to develop an optimal data mixing strategy for the fixed data mixture to achieve the best possible performance on the target model~\citep{xie2023doremi,liu2024regmix,xia2023sheared,luo2024velocitune}. Doremi~\citep{xie2023doremi} trains a reference model based on initial domain weights, which is used to guide the training of another proxy model with the group DRO~\citep{sagawa2019distributionally} algorithm to determine the optimal domain weights for the target model. RegMix~\citep{liu2024regmix} trained large quantities of small proxy models on random
domain distributions, then evaluates these models on the target benchmarks. The best
data mixing recipe is determined by fitting a regression model to these data and selecting distributions that lead to the highest scores. Other works focus on balancing the loss of multiple target fields to achieve balanced optimization~\citep{xia2023sheared, luo2024velocitune}. For example, \citet{xia2023sheared} proposed a batch loading algorithm that loads training data from each domain in proportion to its corresponding rate of loss reduction, which increases the future domain distributions for domains that have slow loss reduction.

Recent works have also explored the effect of domain space definition on data reweighting performance~\citep{wettig2025organize,rukhovich2025commute,diao2025climb,xi2025samplemix}, strengthening the importance of carefully defined domains. For example, previous data mixing methods mostly utilized the default domain space defined by data sources. \citet{wettig2025organize} carefully defined a 24-dimensional domain space from both the topic (e.g., Science$\&$Tech, Fashion$\&$Beauty) and format (e.g., Academic writing, Content listing) perspectives, and re-organized the training data into these domain spaces. Extensive data mixing experiments on these novel domain spaces showed their effectiveness in improving model training performances compared to the source-based domain space. Inspired by their success, we also train the Data Mixing Agent based on these superior ways of domain space definition.

%% file: references.bib
@article{yang2025qwen3,
  title={Qwen3 technical report},
  author={Yang, An and Li, Anfeng and Yang, Baosong and Zhang, Beichen and Hui, Binyuan and Zheng, Bo and Yu, Bowen and Gao, Chang and Huang, Chengen and Lv, Chenxu and others},
  journal={arXiv preprint arXiv:2505.09388},
  year={2025}
}

@inproceedings{du2022glam,
  title={Glam: Efficient scaling of language models with mixture-of-experts},
  author={Du, Nan and Huang, Yanping and Dai, Andrew M and Tong, Simon and Lepikhin, Dmitry and Xu, Yuanzhong and Krikun, Maxim and Zhou, Yanqi and Yu, Adams Wei and Firat, Orhan and others},
  booktitle={International conference on machine learning},
  pages={5547--5569},
  year={2022},
  organization={PMLR}
}

@article{gao2020pile,
  title={The pile: An 800gb dataset of diverse text for language modeling},
  author={Gao, Leo and Biderman, Stella and Black, Sid and Golding, Laurence and Hoppe, Travis and Foster, Charles and Phang, Jason and He, Horace and Thite, Anish and Nabeshima, Noa and others},
  journal={arXiv preprint arXiv:2101.00027},
  year={2020}
}

@inproceedings{biderman2023pythia,
  title={Pythia: A suite for analyzing large language models across training and scaling},
  author={Biderman, Stella and Schoelkopf, Hailey and Anthony, Quentin Gregory and Bradley, Herbie and O’Brien, Kyle and Hallahan, Eric and Khan, Mohammad Aflah and Purohit, Shivanshu and Prashanth, USVSN Sai and Raff, Edward and others},
  booktitle={International Conference on Machine Learning},
  pages={2397--2430},
  year={2023},
  organization={PMLR}
}

@article{zellers2019hellaswag,
  title={Hellaswag: Can a machine really finish your sentence?},
  author={Zellers, Rowan and Holtzman, Ari and Bisk, Yonatan and Farhadi, Ali and Choi, Yejin},
  journal={arXiv preprint arXiv:1905.07830},
  year={2019}
}

@article{welbl2017crowdsourcing,
  title={Crowdsourcing multiple choice science questions},
  author={Welbl, Johannes and Liu, Nelson F and Gardner, Matt},
  journal={arXiv preprint arXiv:1707.06209},
  year={2017}
}

@article{amini2019mathqa,
  title={Mathqa: Towards interpretable math word problem solving with operation-based formalisms},
  author={Amini, Aida and Gabriel, Saadia and Lin, Peter and Koncel-Kedziorski, Rik and Choi, Yejin and Hajishirzi, Hannaneh},
  journal={arXiv preprint arXiv:1905.13319},
  year={2019}
}

@article{lewkowycz2022solving,
  title={Solving quantitative reasoning problems with language models},
  author={Lewkowycz, Aitor and Andreassen, Anders and Dohan, David and Dyer, Ethan and Michalewski, Henryk and Ramasesh, Vinay and Slone, Ambrose and Anil, Cem and Schlag, Imanol and Gutman-Solo, Theo and others},
  journal={Advances in Neural Information Processing Systems},
  volume={35},
  pages={3843--3857},
  year={2022}
}

@article{hu2024openrlhf,
  title={Openrlhf: An easy-to-use, scalable and high-performance rlhf framework},
  author={Hu, Jian and Wu, Xibin and Zhu, Zilin and Wang, Weixun and Zhang, Dehao and Cao, Yu and others},
  journal={arXiv preprint arXiv:2405.11143},
  year={2024}
}

@article{d3rlpy,
  author  = {Takuma Seno and Michita Imai},
  title   = {d3rlpy: An Offline Deep Reinforcement Learning Library},
  journal = {Journal of Machine Learning Research},
  year    = {2022},
  volume  = {23},
  number  = {315},
  pages   = {1--20},
  url     = {http://jmlr.org/papers/v23/22-0017.html}
}

@article{shoeybi2019megatron,
  title={Megatron-lm: Training multi-billion parameter language models using model parallelism},
  author={Shoeybi, Mohammad and Patwary, Mostofa and Puri, Raul and LeGresley, Patrick and Casper, Jared and Catanzaro, Bryan},
  journal={arXiv preprint arXiv:1909.08053},
  year={2019}
}

@article{cobbe2021training,
  title={Training verifiers to solve math word problems},
  author={Cobbe, Karl and Kosaraju, Vineet and Bavarian, Mohammad and Chen, Mark and Jun, Heewoo and Kaiser, Lukasz and Plappert, Matthias and Tworek, Jerry and Hilton, Jacob and Nakano, Reiichiro and others},
  journal={arXiv preprint arXiv:2110.14168},
  year={2021}
}

@article{liu2020logiqa,
  title={Logiqa: A challenge dataset for machine reading comprehension with logical reasoning},
  author={Liu, Jian and Cui, Leyang and Liu, Hanmeng and Huang, Dandan and Wang, Yile and Zhang, Yue},
  journal={arXiv preprint arXiv:2007.08124},
  year={2020}
}

@inproceedings{bisk2020piqa,
  title={Piqa: Reasoning about physical commonsense in natural language},
  author={Bisk, Yonatan and Zellers, Rowan and Gao, Jianfeng and Choi, Yejin and others},
  booktitle={Proceedings of the AAAI conference on artificial intelligence},
  volume={34},
  number={05},
  pages={7432--7439},
  year={2020}
}

@article{clark2018think,
  title={Think you have solved question answering? try arc, the ai2 reasoning challenge},
  author={Clark, Peter and Cowhey, Isaac and Etzioni, Oren and Khot, Tushar and Sabharwal, Ashish and Schoenick, Carissa and Tafjord, Oyvind},
  journal={arXiv preprint arXiv:1803.05457},
  year={2018}
}

@article{sakaguchi2021winogrande,
  title={Winogrande: An adversarial winograd schema challenge at scale},
  author={Sakaguchi, Keisuke and Bras, Ronan Le and Bhagavatula, Chandra and Choi, Yejin},
  journal={Communications of the ACM},
  volume={64},
  number={9},
  pages={99--106},
  year={2021},
  publisher={ACM New York, NY, USA}
}

@article{mihaylov2018can,
  title={Can a suit of armor conduct electricity? a new dataset for open book question answering},
  author={Mihaylov, Todor and Clark, Peter and Khot, Tushar and Sabharwal, Ashish},
  journal={arXiv preprint arXiv:1809.02789},
  year={2018}
}

@article{yang2024qwen2,
  title={Qwen2. 5-math technical report: Toward mathematical expert model via self-improvement},
  author={Yang, An and Zhang, Beichen and Hui, Binyuan and Gao, Bofei and Yu, Bowen and Li, Chengpeng and Liu, Dayiheng and Tu, Jianhong and Zhou, Jingren and Lin, Junyang and others},
  journal={arXiv preprint arXiv:2409.12122},
  year={2024}
}

@article{shao2024deepseekmath,
  title={Deepseekmath: Pushing the limits of mathematical reasoning in open language models},
  author={Shao, Zhihong and Wang, Peiyi and Zhu, Qihao and Xu, Runxin and Song, Junxiao and Bi, Xiao and Zhang, Haowei and Zhang, Mingchuan and Li, YK and Wu, Y and others},
  journal={arXiv preprint arXiv:2402.03300},
  year={2024}
}

@article{hui2024qwen2,
  title={Qwen2. 5-coder technical report},
  author={Hui, Binyuan and Yang, Jian and Cui, Zeyu and Yang, Jiaxi and Liu, Dayiheng and Zhang, Lei and Liu, Tianyu and Zhang, Jiajun and Yu, Bowen and Lu, Keming and others},
  journal={arXiv preprint arXiv:2409.12186},
  year={2024}
}

@article{guo2024deepseek,
  title={DeepSeek-Coder: When the Large Language Model Meets Programming--The Rise of Code Intelligence},
  author={Guo, Daya and Zhu, Qihao and Yang, Dejian and Xie, Zhenda and Dong, Kai and Zhang, Wentao and Chen, Guanting and Bi, Xiao and Wu, Yu and Li, YK and others},
  journal={arXiv preprint arXiv:2401.14196},
  year={2024}
}

@article{liu2024deepseek,
  title={Deepseek-v3 technical report},
  author={Liu, Aixin and Feng, Bei and Xue, Bing and Wang, Bingxuan and Wu, Bochao and Lu, Chengda and Zhao, Chenggang and Deng, Chengqi and Zhang, Chenyu and Ruan, Chong and others},
  journal={arXiv preprint arXiv:2412.19437},
  year={2024}
}

@article{hendrycks2020measuring,
  title={Measuring massive multitask language understanding},
  author={Hendrycks, Dan and Burns, Collin and Basart, Steven and Zou, Andy and Mazeika, Mantas and Song, Dawn and Steinhardt, Jacob},
  journal={arXiv preprint arXiv:2009.03300},
  year={2020}
}

@article{chen2021evaluating,
  title={Evaluating large language models trained on code},
  author={Chen, Mark and Tworek, Jerry and Jun, Heewoo and Yuan, Qiming and Pinto, Henrique Ponde De Oliveira and Kaplan, Jared and Edwards, Harri and Burda, Yuri and Joseph, Nicholas and Brockman, Greg and others},
  journal={arXiv preprint arXiv:2107.03374},
  year={2021}
}

@article{shen2023slimpajama,
  title={Slimpajama-dc: Understanding data combinations for llm training},
  author={Shen, Zhiqiang and Tao, Tianhua and Ma, Liqun and Neiswanger, Willie and Liu, Zhengzhong and Wang, Hongyi and Tan, Bowen and Hestness, Joel and Vassilieva, Natalia and Soboleva, Daria and others},
  journal={arXiv preprint arXiv:2309.10818},
  year={2023}
}

@article{sagawa2019distributionally,
  title={Distributionally robust neural networks for group shifts: On the importance of regularization for worst-case generalization},
  author={Sagawa, Shiori and Koh, Pang Wei and Hashimoto, Tatsunori B and Liang, Percy},
  journal={arXiv preprint arXiv:1911.08731},
  year={2019}
}

@article{luo2023empirical,
  title={An empirical study of catastrophic forgetting in large language models during continual fine-tuning},
  author={Luo, Yun and Yang, Zhen and Meng, Fandong and Li, Yafu and Zhou, Jie and Zhang, Yue},
  journal={arXiv preprint arXiv:2308.08747},
  year={2023}
}

@article{tu2024towards,
  title={Towards generalist biomedical AI},
  author={Tu, Tao and Azizi, Shekoofeh and Driess, Danny and Schaekermann, Mike and Amin, Mohamed and Chang, Pi-Chuan and Carroll, Andrew and Lau, Charles and Tanno, Ryutaro and Ktena, Ira and others},
  journal={Nejm Ai},
  volume={1},
  number={3},
  pages={AIoa2300138},
  year={2024},
  publisher={Massachusetts Medical Society}
}

@article{xie2024finben,
  title={Finben: A holistic financial benchmark for large language models},
  author={Xie, Qianqian and Han, Weiguang and Chen, Zhengyu and Xiang, Ruoyu and Zhang, Xiao and He, Yueru and Xiao, Mengxi and Li, Dong and Dai, Yongfu and Feng, Duanyu and others},
  journal={Advances in Neural Information Processing Systems},
  volume={37},
  pages={95716--95743},
  year={2024}
}

@article{yu2023metamath,
  title={Metamath: Bootstrap your own mathematical questions for large language models},
  author={Yu, Longhui and Jiang, Weisen and Shi, Han and Yu, Jincheng and Liu, Zhengying and Zhang, Yu and Kwok, James T and Li, Zhenguo and Weller, Adrian and Liu, Weiyang},
  journal={arXiv preprint arXiv:2309.12284},
  year={2023}
}

@article{wang2023mathcoder,
  title={Mathcoder: Seamless code integration in llms for enhanced mathematical reasoning},
  author={Wang, Ke and Ren, Houxing and Zhou, Aojun and Lu, Zimu and Luo, Sichun and Shi, Weikang and Zhang, Renrui and Song, Linqi and Zhan, Mingjie and Li, Hongsheng},
  journal={arXiv preprint arXiv:2310.03731},
  year={2023}
}

@article{ivison2024unpacking,
  title={Unpacking dpo and ppo: Disentangling best practices for learning from preference feedback},
  author={Ivison, Hamish and Wang, Yizhong and Liu, Jiacheng and Wu, Zeqiu and Pyatkin, Valentina and Lambert, Nathan and Smith, Noah A and Choi, Yejin and Hajishirzi, Hanna},
  journal={Advances in neural information processing systems},
  volume={37},
  pages={36602--36633},
  year={2024}
}

@article{austin2021program,
  title={Program synthesis with large language models},
  author={Austin, Jacob and Odena, Augustus and Nye, Maxwell and Bosma, Maarten and Michalewski, Henryk and Dohan, David and Jiang, Ellen and Cai, Carrie and Terry, Michael and Le, Quoc and others},
  journal={arXiv preprint arXiv:2108.07732},
  year={2021}
}

@article{xie2023doremi,
  title={Doremi: Optimizing data mixtures speeds up language model pretraining},
  author={Xie, Sang Michael and Pham, Hieu and Dong, Xuanyi and Du, Nan and Liu, Hanxiao and Lu, Yifeng and Liang, Percy S and Le, Quoc V and Ma, Tengyu and Yu, Adams Wei},
  journal={Advances in Neural Information Processing Systems},
  volume={36},
  pages={69798--69818},
  year={2023}
}

@article{wettig2025organize,
  title={Organize the Web: Constructing Domains Enhances Pre-Training Data Curation},
  author={Wettig, Alexander and Lo, Kyle and Min, Sewon and Hajishirzi, Hannaneh and Chen, Danqi and Soldaini, Luca},
  journal={arXiv preprint arXiv:2502.10341},
  year={2025}
}

@article{diao2025climb,
  title={CLIMB: CLustering-based Iterative Data Mixture Bootstrapping for Language Model Pre-training},
  author={Diao, Shizhe and Yang, Yu and Fu, Yonggan and Dong, Xin and Su, Dan and Kliegl, Markus and Chen, Zijia and Belcak, Peter and Suhara, Yoshi and Yin, Hongxu and others},
  journal={arXiv preprint arXiv:2504.13161},
  year={2025}
}

@article{xi2025samplemix,
  title={SampleMix: A Sample-wise Pre-training Data Mixing Strategey by Coordinating Data Quality and Diversity},
  author={Xi, Xiangyu and Kong, Deyang and Yang, Jian and Yang, Jiawei and Chen, Zhengyu and Wang, Wei and Wang, Jingang and Cai, Xunliang and Zhang, Shikun and Ye, Wei},
  journal={arXiv preprint arXiv:2503.01506},
  year={2025}
}

@article{rukhovich2025commute,
  title={Commute Your Domains: Trajectory Optimality Criterion for Multi-Domain Learning},
  author={Rukhovich, Alexey and Podolskiy, Alexander and Piontkovskaya, Irina},
  journal={arXiv preprint arXiv:2501.15556},
  year={2025}
}

@article{lin2025sigma,
  title={Sigma: Differential Rescaling of Query, Key and Value for Efficient Language Models},
  author={Lin, Zhenghao and Tang, Zihao and Liu, Xiao and Gong, Yeyun and Cheng, Yi and Chen, Qi and Li, Hang and Xin, Ying and Yang, Ziyue and Yang, Kailai and others},
  journal={arXiv preprint arXiv:2501.13629},
  year={2025}
}

@article{shi2024continual,
  title={Continual learning of large language models: A comprehensive survey},
  author={Shi, Haizhou and Xu, Zihao and Wang, Hengyi and Qin, Weiyi and Wang, Wenyuan and Wang, Yibin and Wang, Zifeng and Ebrahimi, Sayna and Wang, Hao},
  journal={ACM Computing Surveys},
  year={2024},
  publisher={ACM New York, NY}
}

@article{olmo20242,
  title={2 OLMo 2 Furious},
  author={OLMo, Team and Walsh, Pete and Soldaini, Luca and Groeneveld, Dirk and Lo, Kyle and Arora, Shane and Bhagia, Akshita and Gu, Yuling and Huang, Shengyi and Jordan, Matt and others},
  journal={arXiv preprint arXiv:2501.00656},
  year={2024}
}

@article{hoffmann2022training,
  title={Training compute-optimal large language models},
  author={Hoffmann, Jordan and Borgeaud, Sebastian and Mensch, Arthur and Buchatskaya, Elena and Cai, Trevor and Rutherford, Eliza and Casas, Diego de Las and Hendricks, Lisa Anne and Welbl, Johannes and Clark, Aidan and others},
  journal={arXiv preprint arXiv:2203.15556},
  year={2022}
}

@article{penedo2024fineweb,
  title={The fineweb datasets: Decanting the web for the finest text data at scale},
  author={Penedo, Guilherme and Kydl{\'\i}{\v{c}}ek, Hynek and Lozhkov, Anton and Mitchell, Margaret and Raffel, Colin A and Von Werra, Leandro and Wolf, Thomas and others},
  journal={Advances in Neural Information Processing Systems},
  volume={37},
  pages={30811--30849},
  year={2024}
}

@article{su2024nemotron,
  title={Nemotron-CC: Transforming Common Crawl into a Refined Long-Horizon Pretraining Dataset},
  author={Su, Dan and Kong, Kezhi and Lin, Ying and Jennings, Joseph and Norick, Brandon and Kliegl, Markus and Patwary, Mostofa and Shoeybi, Mohammad and Catanzaro, Bryan},
  journal={arXiv preprint arXiv:2412.02595},
  year={2024}
}

@article{sutton1999policy,
  title={Policy gradient methods for reinforcement learning with function approximation},
  author={Sutton, Richard S and McAllester, David and Singh, Satinder and Mansour, Yishay},
  journal={Advances in neural information processing systems},
  volume={12},
  year={1999}
}

@article{li2025mira,
  title={MIRA: Medical Time Series Foundation Model for Real-World Health Data},
  author={Li, Hao and Deng, Bowen and Xu, Chang and Feng, Zhiyuan and Schlegel, Viktor and Huang, Yu-Hao and Sun, Yizheng and Sun, Jingyuan and Yang, Kailai and Yu, Yiyao and others},
  journal={arXiv preprint arXiv:2506.07584},
  year={2025}
}

@article{zhang2024large,
  title={Large language models for time series: A survey},
  author={Zhang, Xiyuan and Chowdhury, Ranak Roy and Gupta, Rajesh K and Shang, Jingbo},
  journal={arXiv preprint arXiv:2402.01801},
  year={2024}
}

@article{vaswani2017attention,
  title={Attention is all you need},
  author={Vaswani, Ashish and Shazeer, Noam and Parmar, Niki and Uszkoreit, Jakob and Jones, Llion and Gomez, Aidan N and Kaiser, {\L}ukasz and Polosukhin, Illia},
  journal={Advances in neural information processing systems},
  volume={30},
  year={2017}
}

@article{luo2024velocitune,
  title={Velocitune: A Velocity-based Dynamic Domain Reweighting Method for Continual Pre-training},
  author={Luo, Zheheng and Zhang, Xin and Liu, Xiao and Li, Haoling and Gong, Yeyun and Qi, Chen and Cheng, Peng},
  journal={arXiv preprint arXiv:2411.14318},
  year={2024}
}

@article{xia2023sheared,
  title={Sheared llama: Accelerating language model pre-training via structured pruning},
  author={Xia, Mengzhou and Gao, Tianyu and Zeng, Zhiyuan and Chen, Danqi},
  journal={arXiv preprint arXiv:2310.06694},
  year={2023}
}

@article{li2024datacomp,
  title={Datacomp-lm: In search of the next generation of training sets for language models},
  author={Li, Jeffrey and Fang, Alex and Smyrnis, Georgios and Ivgi, Maor and Jordan, Matt and Gadre, Samir Yitzhak and Bansal, Hritik and Guha, Etash and Keh, Sedrick Scott and Arora, Kushal and others},
  journal={Advances in Neural Information Processing Systems},
  volume={37},
  pages={14200--14282},
  year={2024}
}

@article{kumar2020conservative,
  title={Conservative q-learning for offline reinforcement learning},
  author={Kumar, Aviral and Zhou, Aurick and Tucker, George and Levine, Sergey},
  journal={Advances in neural information processing systems},
  volume={33},
  pages={1179--1191},
  year={2020}
}

@article{hendrycks2021measuring,
  title={Measuring mathematical problem solving with the math dataset},
  author={Hendrycks, Dan and Burns, Collin and Kadavath, Saurav and Arora, Akul and Basart, Steven and Tang, Eric and Song, Dawn and Steinhardt, Jacob},
  journal={arXiv preprint arXiv:2103.03874},
  year={2021}
}

@article{grattafiori2024llama,
  title={The llama 3 herd of models},
  author={Grattafiori, Aaron and Dubey, Abhimanyu and Jauhri, Abhinav and Pandey, Abhinav and Kadian, Abhishek and Al-Dahle, Ahmad and Letman, Aiesha and Mathur, Akhil and Schelten, Alan and Vaughan, Alex and others},
  journal={arXiv e-prints},
  pages={arXiv--2407},
  year={2024}
}

@article{liu2024regmix,
  title={Regmix: Data mixture as regression for language model pre-training},
  author={Liu, Qian and Zheng, Xiaosen and Muennighoff, Niklas and Zeng, Guangtao and Dou, Longxu and Pang, Tianyu and Jiang, Jing and Lin, Min},
  journal={arXiv preprint arXiv:2407.01492},
  year={2024}
}

@article{ye2024data,
  title={Data mixing laws: Optimizing data mixtures by predicting language modeling performance},
  author={Ye, Jiasheng and Liu, Peiju and Sun, Tianxiang and Zhan, Jun and Zhou, Yunhua and Qiu, Xipeng},
  journal={arXiv preprint arXiv:2403.16952},
  year={2024}
}
